%% file: main.tex
\documentclass[11pt]{article}
\usepackage[final]{neurips_2025}
\input{macro}

\title{\GraSS{}\includegraphics[height=1em]{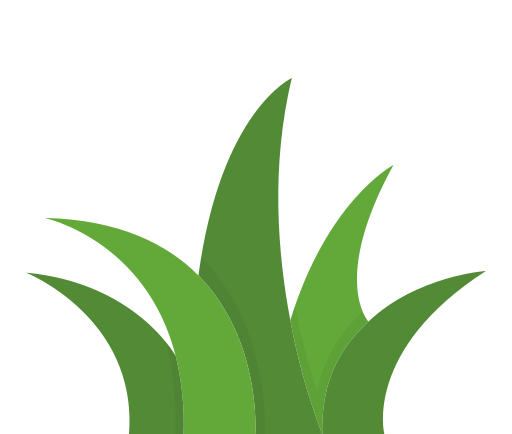}: Scalable Data Attribution with\\ Gradient Sparsification and Sparse Projection}

\author{%
  Pingbang Hu\textsuperscript{\textnormal{1}}\quad
  Joseph Melkonian\textsuperscript{\textnormal{2}}\quad
  Weijing Tang\textsuperscript{\textnormal{3}}\quad
  Han Zhao\textsuperscript{\textnormal{1}}\quad
  Jiaqi W.\ Ma\textsuperscript{\textnormal{1}}\\
  \textsuperscript{1}University of Illinois Urbana-Champaign \quad
  \textsuperscript{2}Womp Labs \quad
  \textsuperscript{3}Carnegie Mellon University\\
  {\small
      \texttt{\{pbb,hanzhao,jiaqima\}@illinois.edu}\quad
      \texttt{joe@womplabs.ai}\quad
      \texttt{weijingt@andrew.cmu.edu}
  }
}

\begin{document}

\maketitle

\input{0_abstract}
\input{1_intro}
\input{2_prelim}
\input{3_compression}
\input{4_experiment}
\input{5_conclusion}

\input{acknowledgement}

\section*{Broader impacts}
This paper presents work whose goal is to advance the efficiency of data attribution algorithms. As there are many socially important applications of data attribution, with the fact that the present SOTA reliable data attribution methods do not scale well to commercial-size LLMs, there are many potential societal consequences of our work. However, none of which we feel must be specifically highlighted here.

\bibliography{reference}
\bibliographystyle{abbrvnat}

\newpage
\appendix
\input{6_appendix}

\end{document}

%% file: macro.tex
\usepackage{amsmath,amsfonts,amssymb,amsthm,commath,dsfont,nicefrac,xfrac,mathtools,mathrsfs}
\usepackage[T1]{fontenc}
\usepackage{float}
\usepackage{bm,bbm}
\usepackage[inline]{enumitem}
\usepackage{framed}
\usepackage{xspace}
\usepackage{microtype}
\usepackage[dvipsnames]{xcolor}
\usepackage[toc,page]{appendix}
\usepackage{svg}
\usepackage{booktabs}       
\usepackage{icomma}
\usepackage{hyperref}
\hypersetup{
    colorlinks=true,
    linkcolor=black,
    citecolor=blue,
    filecolor=magenta,
    urlcolor=blue
}
\usepackage{cleveref}

\usepackage{caption}
\usepackage{subcaption}
\Crefname{equation}{Eq.\!}{Eqs.\!}
\Crefname{subtable}{Table}{Tables}

\theoremstyle{plain}
\newtheorem{theorem}{Theorem}[section]

\theoremstyle{plain}

\theoremstyle{plain}
\newtheorem{remark}[theorem]{Remark}

\usepackage{graphicx}
\usepackage{wrapfig}

\renewcommand{\parallel}{\mathrel{/\mkern-5mu/}}
\makeatletter
\newcommand{\notparallel}{%
    \mathrel{\mathpalette\not@parallel\relax}%
}
\newcommand{\not@parallel}[2]{%
    \ooalign{\reflectbox{\(\m@th#1\smallsetminus\)}\cr\hfil\(\m@th#1\parallel\)\cr}%
}
\makeatother

\DeclareMathOperator*{\argmax}{arg\,max}
\DeclareMathOperator*{\argmin}{arg\,min}
\DeclareMathOperator{\diag}{diag}
\DeclareMathOperator{\RM}{\textnormal{\textsc{RM}}}
\DeclareMathOperator{\SM}{\textnormal{\textsc{SM}}}
\DeclareMathOperator{\Mask}{\textnormal{\textsc{Mask}}}
\DeclareMathOperator{\FJLT}{\textnormal{\textsc{FJLT}}}
\DeclareMathOperator{\SJLT}{\textnormal{\textsc{SJLT}}}
\DeclareMathOperator{\Gauss}{\textnormal{\textsc{Gauss}}}

\newcommand{\Trak}{\textsc{Trak}}

\newcommand{\GraSS}{\textsc{GraSS}}
\newcommand{\FactGraSS}{\textsc{FactGraSS}}
\newcommand{\LoGra}{\textsc{LoGra}}
\newcommand{\Random}{\textsc{Random}}
\newcommand{\GradDot}{\textsc{GradDot}}

\usepackage{titlesec}
\titlespacing*{\section}{0pt}{*1.6}{*0.8}
\titlespacing*{\subsection}{0pt}{*1.5}{*0.6}

\makeatletter
\renewcommand\paragraph{\@startsection{paragraph}{4}{\z@}%
    {0ex \@plus1ex \@minus1ex}%
    {-1em}%
    {\normalfont\normalsize\bfseries}}

\makeatother
\everypar{\looseness=-1}

\usepackage{import}
\usepackage{xifthen}
\usepackage{pdfpages}
\usepackage{transparent}

\usepackage{tcolorbox}

\definecolor{lightgreen}{RGB}{232, 245, 233}
\definecolor{lightred}{RGB}{255, 235, 205}
\definecolor{dreamcolor}{RGB}{217, 128, 110}
\definecolor{controlcolor}{RGB}{217, 128, 110}

%% file: 0_abstract.tex
\renewcommand*{\thefootnote}{\fnsymbol{footnote}}
\begin{abstract}
    Gradient-based data attribution methods, such as influence functions, are critical for understanding the impact of individual training samples without requiring repeated model retraining. However, their scalability is often limited by the high computational and memory costs associated with per-sample gradient computation. In this work, we propose \GraSS{}, a novel gradient compression algorithm and its variants \FactGraSS{} for linear layers specifically, that explicitly leverage the inherent sparsity of per-sample gradients to achieve sub-linear space and time complexity. Extensive experiments demonstrate the effectiveness of our approach, achieving substantial speedups while preserving data influence fidelity. In particular, \FactGraSS{} achieves up to \(165\%\) faster throughput on billion-scale models compared to the previous state-of-the-art baselines.\footnote[2]{Our code is publicly available at \url{https://github.com/TRAIS-Lab/GraSS}.}
\end{abstract}
\renewcommand*{\thefootnote}{\arabic{footnote}}
\setcounter{footnote}{0}

%% file: 1_intro.tex
\section{Introduction}\label{sec:introduction}
Data attribution~\citep{deng2025survey} aims to measure the impact of individual training samples on a machine learning model and has been widely applied to data-centric problems in modern AI, such as data curation~\citep{koh2017understanding}, fact tracing~\citep{lin2024token}, and data compensation~\citep{deng2024computationalcopyrightroyaltymodel}. There are two major categories of data attribution methods, \emph{gradient-based} and \emph{retraining-based}~\citep{hammoudeh2024training}. The former category, such as influence functions~\citep{koh2017understanding} and its variants, has gained increasing popularity in large-scale applications as it does not require costly model retraining. One common feature of gradient-based methods is their reliance on the \emph{per-sample gradient}---the gradient of the loss with respect to model parameters for each individual data point---to capture the local sensitivity of the model with respect to each training sample, providing a fine-grained understanding of data influence.

However, gradient-based methods still face significant scalability challenges for very large models, such as large language models (LLMs). Specifically, computing and storing per-sample gradients for a model with \(n\) training samples and \(p\) parameters requires \(O(np)\) memory and compute, creating a severe bottleneck for large-scale models. To address this, recent work has explored compressing these high-dimensional gradients into lower-dimensional representations, reducing memory requirements to \(O(nk)\) for a target compression dimension \(k \ll p\)~\citep{wojnowicz2016influence,park2023trak,choe2024your}. However, this compression often introduces additional computational overhead, as the most common approach---random matrix projection with the Johnson-Lindenstrauss (JL) guarantee---requires dense matrix multiplications, resulting in an overall time complexity of \(O(nkp)\).

To overcome these limitations, more specialized approaches have been proposed. For example, the fast Johnson-Lindenstrauss transform (FJLT) used in \Trak{}'s official implementation~\citep{park2023trak} exploits structured random matrices, reducing the projection time to \(O((p+k) \log p)\) per sample. Alternatively, recent work by \citet{choe2024your} proposed \LoGra{} that leverages the Kronecker product structure of gradients in linear layers, reducing projection time to \(O(\sqrt{pk})\) for each data via a factorized matrix approach. However, these methods are often designed for general inputs and do not fully exploit the unique \emph{sparsity structures} present in per-sample gradients.

In this paper, we push the boundaries of these state-of-the-art (SOTA) gradient compression methods by proposing a novel, two-stage gradient compression algorithm called \GraSS{} (\textbf{Gra}dient \textbf{S}parsification and \textbf{S}parse-projection), that achieves \emph{sub-linear} space and time complexity by explicitly leveraging the inherent sparsity of per-sample gradients. Our contributions are as follows:
\begin{enumerate}[leftmargin=*]
	\item We identify two critical sparsity properties in per-sample gradients and leverage these to develop a gradient compression algorithm, \GraSS{}, that reduces both space and time complexity from \(O(pk)\) to \(O(k^{\prime})\), where \(k^{\prime}\) is a tunable hyperparameter in the range \([k, p]\).

	\item We further derive a practically efficient variant of \GraSS{} for linear layers, \FactGraSS{} (\textbf{Fact}orized \GraSS{}), which exploits the gradient structures for linear layers, similarly achieving a time and space complexity of \(O(k^{\prime})\) \textbf{without} the need to ever materialize the full gradients.

	\item Through extensive experiments, we demonstrate that our approach achieves several orders of magnitude speedup compared to previous SOTA while maintaining competitive performance on standard evaluation metrics. In particular, on billion-scale language models and datasets, \FactGraSS{} is up to \(165\%\) faster in terms of compression throughput compared to \LoGra{}.
\end{enumerate}

%% file: 2_prelim.tex
\section{Preliminary}\label{sec:preliminary}
We begin by introducing the influence function and its practical implementation as our running example of gradient-based data attribution methods. We then demonstrate how random projection can be integrated with the influence function. We note that the proposed compression methods naturally extend to other gradient-based approaches that share similar low-level computations. Examples include TRAK~\citep{park2023trak}, SGD-Influence~\citep{hara2019data}, and Data Value Embedding~\citep{wang2025capturing}. A more detailed discussion of related work is provided in \Cref{adxsubsec:related-work}.

\subsection{Influence function}\label{subsec:influence-function}
Given a dataset \(D = \{z_i \in \mathbb{R}^d\}_{i=1}^n\), consider a model parametrized by \(\hat{\theta} \in \mathbb{R}^p\) that this trained on the dataset \(D\) with a loss function \(\ell \colon \mathbb{R}^{d} \times \mathbb{R}^{p} \to \mathbb{R}\) via empirical risk minimization: \(\hat{\theta} = \argmin_{\theta \in \mathbb{R}^p} \frac{1}{n} \sum_{i=1}^{n} \ell(z_i; \theta)\). Under this setup, the \emph{influence function}~\citep{koh2017understanding} can be theoretically derived; essentially, it gives an estimation on every training data \(z_i\)'s ``influence'' \(\mathcal{I}(z_i, z_{\text{test}})\) of the test loss \(\ell(z_{\text{test}}; \hat{\theta})\) of a given test data \(z_{\text{test}}\) when \(z_i\) is removed from \(D\) as
\[
    \mathcal{I}(z_i, z_{\text{test}})
    \coloneqq \nabla_{\theta } \ell(z_{\text{test}}; \hat{\theta })^{\top} H_{\hat{\theta}}^{-1} \nabla_{\theta } \ell(z_i; \hat{\theta }),
\]
where \(H_{\hat{\theta}} = \frac{1}{n}\sum_{i=1}^{n} \nabla^2_{\theta} \ell(z_i, \hat{\theta})\) is the empirical Hessian. As \(H_{\hat{\theta}} \in \mathbb{R}^{p \times p}\) and computing it requires higher-order differentiation for every training data, several approximation algorithms aim to mitigate this. One famous approximation is the Fisher information matrix (FIM)~\citep{fisher1922mathematical} approximation \(H_{\hat{\theta}} \approx \mathbb{E}_{z} [\nabla_{\theta} \ell(z; \hat{\theta}) \nabla_{\theta} \ell(z; \hat{\theta})^{\top}]\), which is exact for model trained with the negative log-likelihood objective. Since FIM only involves the first-order gradient, which is also needed in the other parts of the calculation of influence \(\mathcal{I}\), and hence is a popular and efficient approximation. Given this, people realize that an efficient way to compute the influence function is to divide the computation into two stages~\citep{lin2024token,choe2024your}:
\begin{enumerate}[leftmargin=*]
    \item \textbf{Cache stage}: \begin{enumerate*}[label=\arabic*.)]
              \item compute all \emph{per-sample gradients} \(g_i \coloneqq g_{z_i} \coloneqq \nabla_{\theta } \ell(z_i; \hat{\theta })\),
              \item construct the FIM \(F_{\hat{\theta } } \coloneqq \frac{1}{n}\sum_{i=1}^{n} g_i g_i^{\top}\),
              \item perform \emph{inverse FIM-vector-product} (iFVP) via \(\widetilde{g} _i \coloneqq F_{\hat{\theta} }^{-1} g_i\) for all \(z_i\)'s.
          \end{enumerate*}
    \item \textbf{Attribute stage}: For a query data \(z_{\text{test} }\), \begin{enumerate*}[label=\arabic*.)]
              \item compute its per-sample gradient \(g_{\text{test} } \coloneqq \nabla_{\theta } \ell(z_{\text{test} }; \hat{\theta })\),
              \item compute all-pair-inner-product between \(g_{\text{test} }\) and \(\{ \widetilde{g} _i \}_{i=1}^{n} \) as \(\mathcal{I} (z_i, z_{\text{test} }) = \langle g_{\text{test} }, \widetilde{g} _i \rangle \) for all \(z_i\)'s.
          \end{enumerate*}
\end{enumerate}
The bottleneck of this pipeline is the cache stage, since the problem for the attribute stage is the well-studied \emph{vector inner product search}, where numerous optimization techniques have been studied in the vector database community. On the other hand, iFVP remains a challenging task due to the matrix inversion of quadratic model size, where in most cases, even materializing FIM is infeasible.

\subsection{Random projection}\label{subsec:random-projection}
Despite multiple attempts to accelerate influence function from various angles, one of the most naive and simple strategies, \Random{}~\citep{wojnowicz2016influence,schioppa2022scaling,park2023trak}, remains practically relevant and achieves SOTA attribution results. \Random{} leverages sketching (random projection) techniques by replacing each per-sample gradient \(g_i \in \mathbb{R} ^{p}\) with \(\hat{g} _i \coloneqq P g_i \in \mathbb{R} ^{k}\), where \(P\in \mathbb{R} ^{k \times p } \) is a random projection matrix for some \(k \ll p\). This subsequently leads to the \emph{projected FIM} approximation \(\hat{F} _{\hat{\theta} } \coloneqq P F_{\hat{\theta } } P^{\top} = \mathbb{E}_{z}[(\hat{g} _z \cdot (\hat{g} _z) ^{\top} )] \in \mathbb{R} ^{k \times k}\), i.e., a restriction of \(F_{\hat{\theta}}\) to the subspace spanned by the columns of \(P\). The theoretical merits of \Random{} largely come from the well-known Johnson-Lindenstrauss lemma~\citep{johnson1984extensions}, which states that for \(P\) drawn appropriately, e.g., \(P_{ij} \overset{\text{i.i.d.} }{\sim } \mathcal{N} (0, 1)\) or \(\mathcal{U} (\{ \pm 1 \} )\) for all \(i, j\),\footnote{We omit the normalization factor (in this case, \(1 / \sqrt{k}\)) to keep the presentation clean.} with high probability, the pair-wise distance \(\lVert g_i - g_j \Vert\) between any two \(g_i\) and \(g_j\) will be preserved up to \(1 \pm \epsilon \) factor after the projection, whenever \(k = O(\epsilon ^{-2} \log n)\). While this does not fully justify whether the inner product between a projected per-test-sample gradient \(\hat{g} _{\text{test} }\) and the ``conditioned'' projected per-train-sample gradient \(\widetilde{\hat{g}} _i \coloneqq (\hat{F} _{\hat{\theta} })^{-1} \hat{g} _i\) will be preserved (see \Cref{adsubsec:a-note-on-Johnson-Lindenstrauss-lemma} for an in-depth discussion), \Random{} remains to be one of the strongest baselines to date and is practically appealing due to its simplicity.\footnote{We follow the same notational convention in the rest of the paper: \(\hat{\cdot} \) denotes compression, \(\widetilde{\cdot} \) denotes (FIM) precondition, and \(\widetilde{\hat{\cdot}}\) denotes the compression and then (compressed FIM) precondition.}

Computational-wise, \Random{} accelerates iFVP significantly as the matrix inversion complexity scales down from \(O(p^2)\) to \(O(k^2)\). In terms of the projection overhead, the matrix-based projection method requires \(O(kp)\) overhead per projection. \Trak{}~\citep{park2023trak} leverages the \emph{fast Johnson-Lindenstrauss transform} (FJLT)~\citep{ailon2009fast,fandina2023fast} that has a similar theoretical guarantee as the random matrix-based projection to achieve a speed up of \(O((p+k) \log p)\). Another line of work by \citet{choe2024your} called \LoGra{} exploits the gradient structure of linear layers and factorizes the projection accordingly, reducing the problem size quadratically. With suitable hyper-parameter choice, the computational complexity goes down from \(O(k_l p_l)\) to \(O(\sqrt{k_l p_l} )\), where \(k_l\) and \(p_l\) now refer to the projection dimension and number of model parameters of one (\(l^{\text{th}}\)) linear layer. This sets the SOTA efficiency and attribution quality to date.

%% file: 3_compression.tex
\section{\GraSS{}: Gradient Sparsification and Sparse projection}\label{sec:GraSS}
In this section, we first explore two key sparsity properties in per-sample gradients (\Cref{subsec:per-sample-gradient-sparsity,subsec:effective-parameter-sparsity}) and propose efficient compression methods for each. Combining them, we present \GraSS{} and its variant \FactGraSS{} (\Cref{subsec:GraSS-FactGraSS-multi-stage-compression}), which beat the previous SOTA data attribution algorithms.

\subsection{Per-sample gradient sparsity}\label{subsec:per-sample-gradient-sparsity}
Modern deep learning models often induce highly sparse per-sample gradients, especially when using popular activation functions like ReLU~\citep{nair2010rectified}. To see this, consider the gradient of the first read-in linear layer with weight \(W \in \mathbb{R}^{d^{\text{out} } \times d^{\text{in} }}\) with \(d^{\text{in} } = d\) and ReLU activations. Then given a sample \(z \in \mathbb{R}^d\), the output is \(h = \operatorname{ReLU}(Wz)\). Since \(\operatorname{ReLU} (x) = \max(0, x)\) sets all negative pre-activations to zero, naturally creating sparse activations.  This sparsity propagates to the gradient computations via the chain rule, resulting in gradients with numerous zero entries. This is not unique to ReLU and extends to many other activation functions that exhibit similar behavior.

\begin{remark}
    Such sparsity is unique to \emph{per-sample} gradients: for mini-batch gradients \(\sum_{i \in B} g_i / \lvert B \rvert \), the sparsity pattern differs for individual \(g_i\) and will be destroyed when adding together.
\end{remark}

Given the inherently sparse nature of these gradients, it is natural to consider other compression methods that can effectively exploit this input sparsity. Traditional dense random projection methods struggle to leverage sparse inputs without incurring significant overhead. Although FJLT is often more efficient, its algorithmic structure also prevents it from effectively exploiting sparsity patterns.

\begin{figure}[thpb]
    \centering
    \begin{minipage}{.48\textwidth}
        \centering
    \def\svgwidth{\columnwidth}
    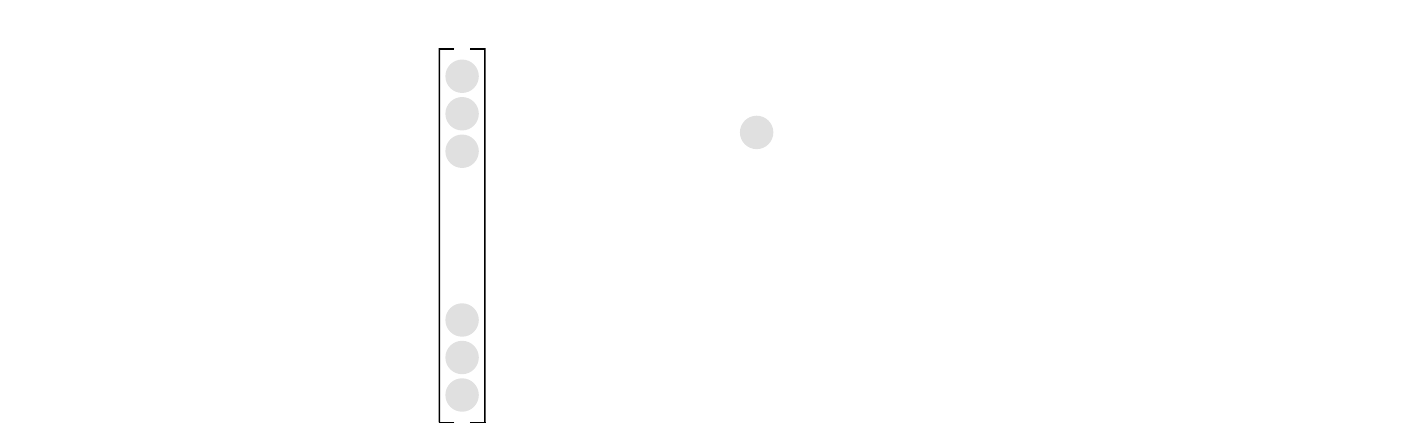

        \caption{Dense Rademacher projection.}
        \label{fig:dense-Rademacher-projection}
    \end{minipage}\hfill%
    \begin{minipage}{.48\textwidth}
        \centering
    \def\svgwidth{\columnwidth}
    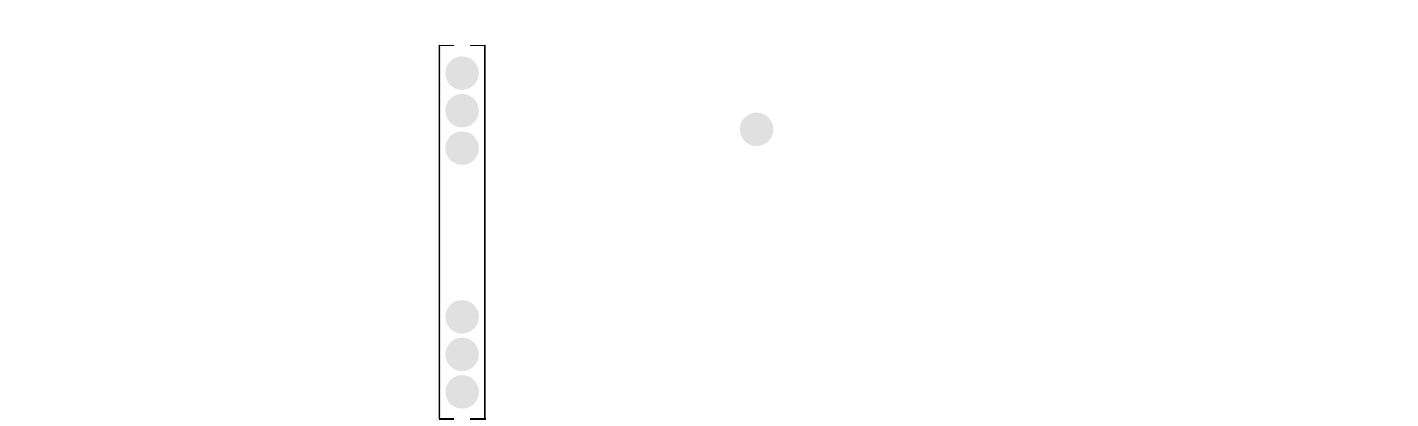

        \caption{Sparse Rademacher projection (\(s = 1\)).}
        \label{fig:sparse-Rademacher-projection}
    \end{minipage}
\end{figure}

\paragraph{Sparse Johnson-Lindenstrauss Transform.}
A natural candidate for efficient gradient projection is the \emph{sparse Johnson-Lindenstrauss transform} (SJLT)~\citep{dasgupta2010sparse,kane2014sparser}, which significantly reduces the computational cost by sparsifying the projection matrix. To understand SJLT, it is useful to revisit the standard matrix-based projection approach, which relies on matrix-vector multiplication. Given a projection matrix \(P \in \mathbb{R}^{k \times p}\) and an input vector \(g \in \mathbb{R}^{p}\) to be projected, the product \(Pg\) can be computed as \(\hat{g} = Pg = \sum_{j=1}^{p} g(j) P_{: j}\), where the \(j^{\text{th}}\) term represents the \(j^{\text{th}}\) column \(P_{: j}\) of \(P\) scaled by the \(j^{\text{th}}\) entry \(g(j)\) of \(g\). In the case of a dense Rademacher projection matrix (entries being \(\pm 1\)), this requires \(O(pk)\) computation for both constructing \(g(j) P_{: j}\) and summing them. This dense projection process is illustrated in \Cref{fig:dense-Rademacher-projection}.

\begin{wrapfigure}[13]{r}{0.32\textwidth}
    \centering
    \vspace{-1\intextsep}
    \def\svgwidth{\columnwidth}
    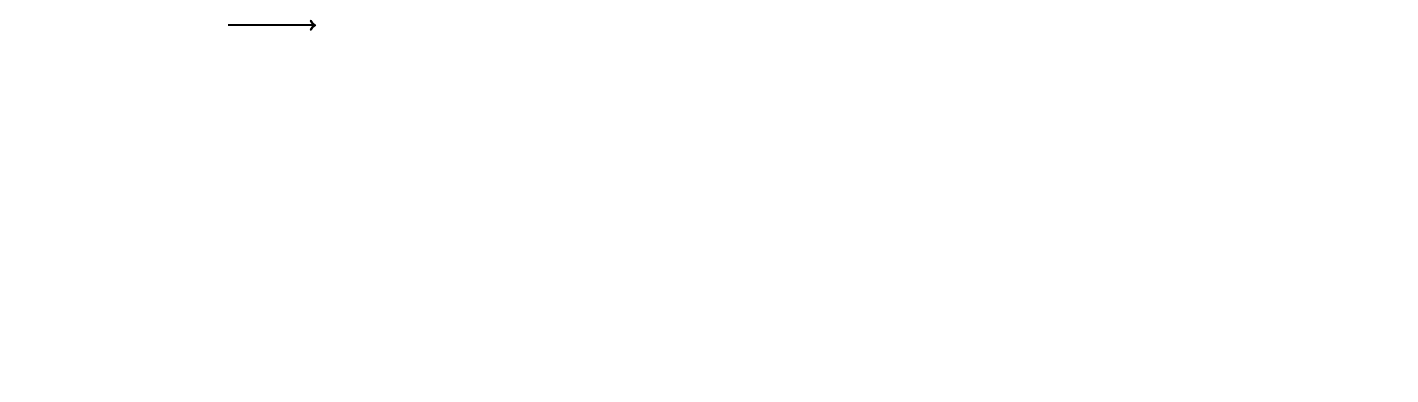

    \caption{SJLT with \(s = 1\).}
    \label{fig:SJLT-computation}
\end{wrapfigure}

From this perspective, the SJLT arises naturally: by ``zeroing'' out the projection matrix \(P\), we significantly reduce the required computation, as illustrated in \Cref{fig:sparse-Rademacher-projection}. Specifically, \citet{dasgupta2010sparse} and \citet{kane2014sparser} demonstrated that retaining only \(s = o_\epsilon(k)\) out of the \(k\) possible non-zero entries for each column of the projection matrix still preserves the essential properties required by the Johnson-Lindenstrauss lemma. This approach, which we denote as \(\SJLT_k(\cdot)\), reduces both the time and space complexity to \(O(ps)\), where \(s = o_\epsilon(k)\) is much smaller than \(k\).

An equivalent way to view the computation of SJLT is shown in \Cref{fig:SJLT-computation}, where we initialize \(\hat{g} \) to be a zero vector and sequentially scan through \(g\), with each \(g(j)\) for \(j \in [p]\) chooses \(s\) many random \(j^{\prime} \in [k]\) to either add on or subtract from the corresponding \(\hat{g} (j^{\prime} )\). It is immediate that if the input is \textbf{itself sparse}, the complexity can be further reduced. Specifically, for a dense matrix projection, the complexity becomes \(O(k \operatorname{nnz}(g))\), and for SJLT, this drops to \(O(s \operatorname{nnz}(g))\), where \(\operatorname{nnz}(g) \coloneqq \lVert g \rVert_0\) denotes the number of non-zero entries in \(g\). We highlight \(g\) in red in \Cref{fig:SJLT-computation} to signify that the computational complexity of SJLT scales with the size of \(g\).

\begin{wrapfigure}[20]{l}{0.32\textwidth}
    \centering
    \vspace{-1\intextsep}
    \includegraphics[width=\linewidth]{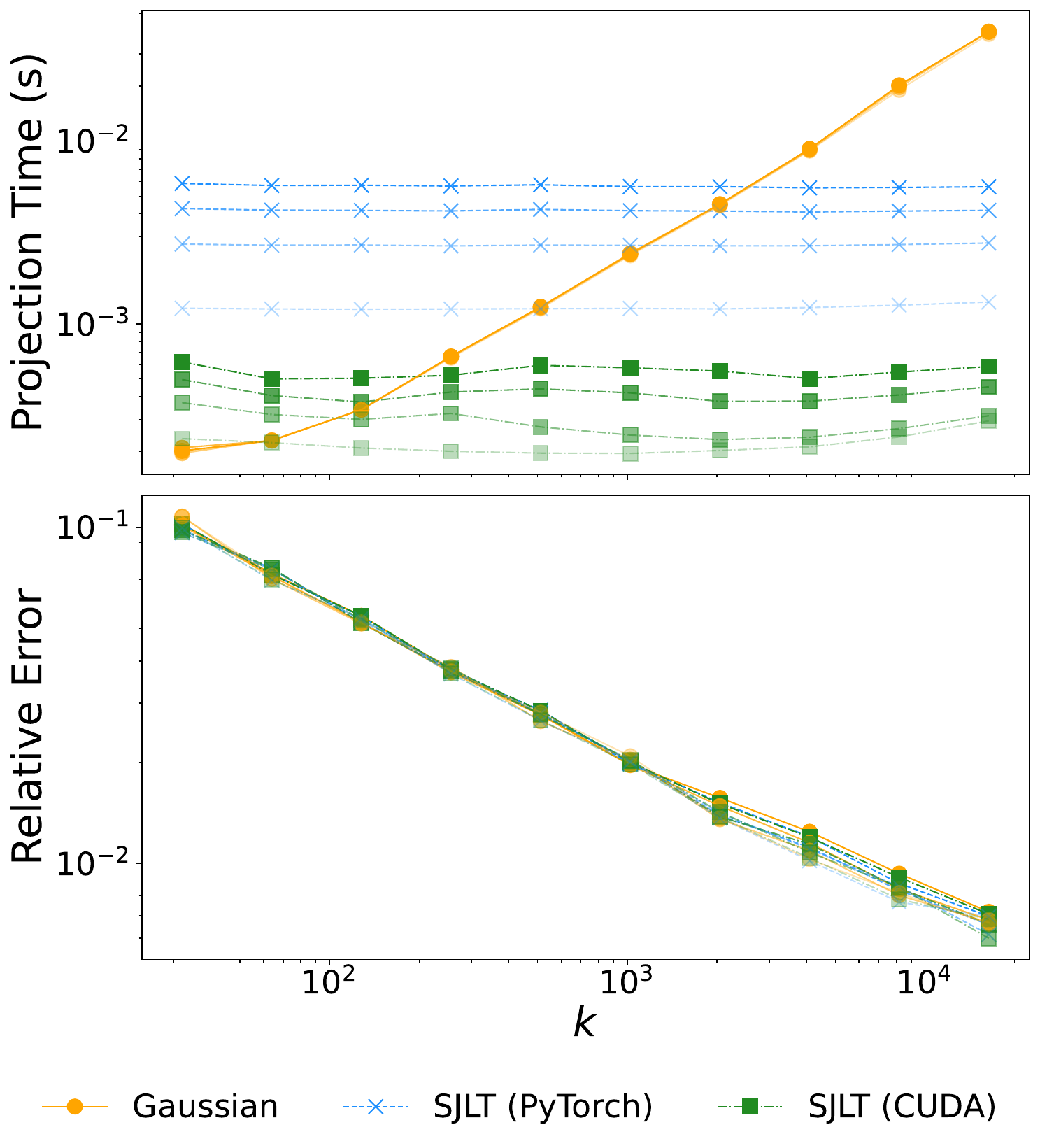}
    \caption{Benchmark of different projection methods with \(p = 131,072\) under several sparsity levels (distinguished by opacity). Relative error is w.r.t.\ pair-wise distance preservation.}
    \label{fig:SJLT-benchmark}
\end{wrapfigure}

\paragraph{Implementation of SJLT.}
Despite these theoretical advantages, practical implementations of SJLT face critical performance challenges, such as thread contention and irregular memory access patterns. While the latter overhead is due to the nature of SJLT, the former occurs because multiple threads may attempt to write to the same entry in the output vector \(\hat{g}\), causing race conditions that degrade performance, especially when the target dimension \(k\) is small. These are especially critical when implementing in general-purpose libraries like \texttt{PyTorch}. Moreover, the default matrix multiplication algorithms in \texttt{PyTorch} are highly hardware-optimized (e.g., cache-friendly memory layouts and fused multiply-accumulate instructions), in practice often outperforming any similar multiplication algorithms when the problem size is small.

To address these issues, we developed a \emph{SJLT CUDA kernel}\footnote{The code is publicly available at \url{https://github.com/TRAIS-Lab/sjlt}.} that optimizes the memory access patterns and minimizes thread contention to better exploit the underlying hardware capabilities. This kernel significantly reduces the overhead compared to its \texttt{PyTorch} implementation counterpart, resulting in substantial performance gains. As shown in \Cref{fig:SJLT-benchmark}, for SJLT with \(s = 1\), our CUDA implementation outperforms the highly optimized dense matrix multiplications for small projection problem sizes, while retaining the speedup of SJLT w.r.t.\ input sparsity. In contrast, dense Gaussian projections exhibit a clear dependency on \(k\) while neglecting the input sparsity, making them less efficient in such cases. In practice, we set \(s = 1\) to optimize for speed while enjoying a strong empirical guarantee for small relative error, as seen in \Cref{fig:SJLT-benchmark}.

To summarize, the complexity of SJLT
\begin{enumerate*}[label=\arabic*.)]
    \item scales with the input sparsity, and
    \item is independent of the target dimension \(k\),
\end{enumerate*}
both of which are critical for efficient gradient compression. Specifically,
\begin{enumerate*}[label=\arabic*.)]
    \item per-sample gradients are naturally sparse, and
    \item larger \(k\) generally improves the fidelity of data attribution, which is more desirable.
\end{enumerate*}
These make SJLT a natural fit for gradient compression.

\subsection{Effective parameter sparsity}\label{subsec:effective-parameter-sparsity}
While SJLT effectively reduces the computational overhead and takes advantage of the sparsity structure of the input, it still scales linearly with the potentially large input dimensionality \(p\). We now explore a more aggressive compression that achieves \emph{sub-linear} complexity by directly exploiting the inherent \emph{effective parameter sparsity} in neural networks. We term this approach as \emph{sparsification}.

\paragraph{Random Mask.}
Modern deep learning models often exhibit a high degree of parameter redundancy, where only a small fraction of the weights significantly contribute to the model’s final performance~\citep{han2015deep,frankle2018the}. In a similar vein, the distributed training community has observed that the majority of gradient updates are redundant, allowing for substantial compression without a significant impact on model accuracy~\citep{lin2018dgc, aji2017sparse}. This suggests that many parameters can be safely ignored without substantial loss in accuracy. Inspired by this observation, a simple yet surprisingly effective sparsification algorithm, \emph{Random Mask}, randomly selects a small subset from the \(p\) input dimensions to form a compressed representation.

\begin{wrapfigure}[13]{r}{0.32\textwidth}
    \centering
    \vspace{-1\intextsep}
    \def\svgwidth{\columnwidth}
    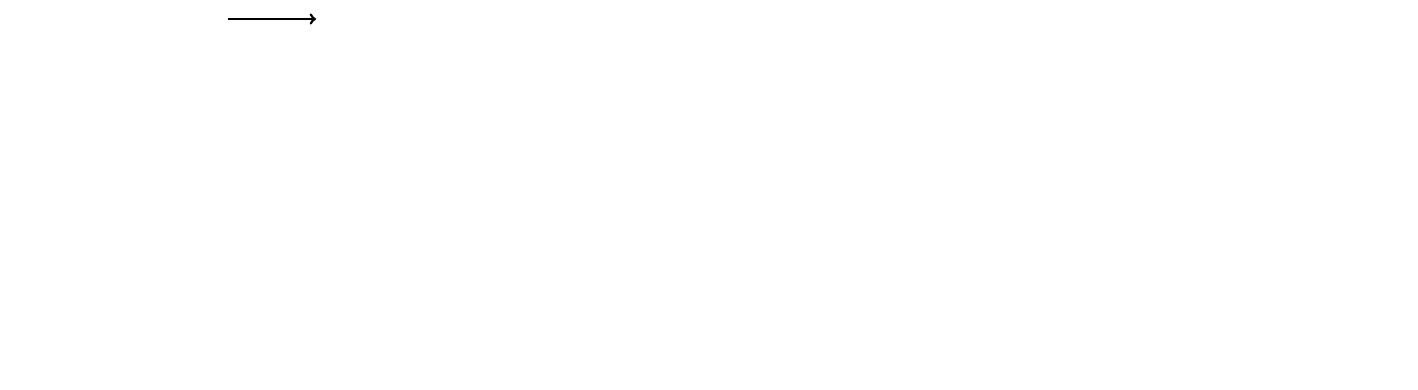

    \caption{Mask.}
    \label{fig:Mask}
\end{wrapfigure}

Formally, the Random Mask (\(\RM_{k}(\cdot)\)) involves selecting a random subset of \(k\) dimensions from the original \(p\)-dimensional gradient vector, effectively extracting a length-\(k\) sub-vector. This can also be viewed as a random projection onto the standard basis of a randomly chosen \(k\)-dimensional Euclidean subspace, i.e., \(\hat{g} = M g\) where \(M \in \mathbb{R}^{k \times p}\) is a sparse binary selection matrix with exactly one (non-repetitive) non-zero entry per row, corresponding to the randomly chosen dimensions. This is illustrated in \Cref{fig:Mask}.

At first glance, \(\RM_k\) may seem overly aggressive, as it discards a substantial amount of information. However, empirical evidence suggests that this method can still yield non-trivial attribution performance, especially when the underlying gradient distribution is sparse or when the model is over-parameterized. Moreover, the extreme simplicity of this approach makes it highly efficient, with a computational cost of just \(O(k)\), achieving a \emph{sub-linear} complexity w.r.t.\ \(p\). We highlight \(\hat{g} \) in red in \Cref{fig:Mask} to signify that the computational complexity of Random Mask scales with the size of \(\hat{g} \).

\paragraph{Selective Mask.}
Building on the idea of Random Mask, we introduce a more structured approach, \emph{Selective Mask} (\(\SM_k(\cdot)\)), which aims to selectively retain the most important parameters based on a simple, yet effective, data-driven optimization. Inspired by recent work on identifying influential model parameters~\citep{he2025localizeandstitch}, Selective Mask introduces a small but meaningful optimization overhead to improve the fidelity of the compressed representation. Formally, given a training set \(\{z_i\}_{i=1}^n\), we define the selective masking problem as the following \emph{unconstrained} optimization task:
\begin{equation}\label{eq:selective-mask}
    S^{\ast}
    = \argmax_{S\in \mathbb{R}^{p}} \mathbb{E}_{z_{\text{test}}} \left[\operatorname{corr}\left( ( \langle g_i , g_{z_{\text{test}}} \rangle )_{i=1}^{n}, ( \langle \hat{g} _i , \hat{g} _{z_{\text{test}}} \rangle )_{i=1}^{n} \right) \right] - \lambda \Vert \sigma(S) \Vert _1,
\end{equation}
where \(\hat{g} _i = \sigma (S) \odot g_i \in \mathbb{R} ^{p}\) is the (soft-)masked \(g_i\), \(\odot\) denotes the element-wise product, and \(\sigma (\cdot ) \in (0, 1)\) is the sigmoid function. The first term of the objective encourages the average correlation between the original and masked gradients' \GradDot{} attribution scores~\citep{charpiat2019input}, a widely used and computationally efficient approximation of the influence function. The second term, an \(\ell _1\) regularization penalty, promotes sparsity by pushing \(\sigma (S)\) towards a binary mask.

Once the optimal \(S^{\ast}\) is obtained after solving \Cref{eq:selective-mask}, the final binary mask \(M \in \{0, 1\}^{k \times p}\) can be extracted by thresholding the sigmoid outputs with \(k = \sum_{j=1}^{p} \mathbbm{1}_{\sigma(S^{\ast})_j \geq 0.5}\) is the number of selected dimensions. Formally, one can obtain the explicit mask matrix \(M\) via solving \(\langle M_{: j} , 1_{k} \rangle = \mathbbm{1}_{\sigma(S^{\ast})_j \geq 0.5}\), but the actual implementation is simply an index extraction for all \(j\) for \(\mathbbm{1}_{\sigma(S^{\ast})_j \geq 0.5}\).

This formulation avoids the exponential complexity of directly optimizing over discrete binary masks, as the continuous nature of \(S\) allows for efficient, first-order gradient-based optimization. While it incurs a one-time overhead for solving \Cref{eq:selective-mask}, this method provides a more principled approach to mask selection by directly targeting a widely used surrogate data attribution score, making it a natural extension of Random Mask. We use \(\Mask_k\) to refer to either \(\RM_k\) or \(\SM_k\) for convenience.

\subsection{\GraSS{} \& \FactGraSS{}: Multi-stage compression}\label{subsec:GraSS-FactGraSS-multi-stage-compression}
We now formally introduce \GraSS{} and \FactGraSS{}, an integration of the proposed approaches by combining the sparse projection (\Cref{subsec:per-sample-gradient-sparsity}) and also sparsification techniques (\Cref{subsec:effective-parameter-sparsity}).

\subsubsection{\GraSS{}: Sparsify first, sparse projection next}\label{subsubsec:GraSS-sparsify-first-sparse-projection-next}
Recall that the time complexity of SJLT with \(s = 1\) is \(O(p)\) where \(p\) is the input dimension, while for both sparsification techniques are \(O(k)\) where \(k\) is the target sparsification dimension. A natural idea is to employ a two-stage compression: Given an input \(g\) and target compression dimension \(k \ll p\),
\begin{enumerate}[leftmargin=*]
    \item \textbf{\emph{Sparsification}}: sparsify the input \(g\) to a sub-vector \(g^{\prime} \) of dimension \(k^{\prime} \) with \(k < k^{\prime} \ll p\).
    \item \textbf{\emph{Sparse projection}}: then apply SJLT to \(g^{\prime} \) to get the compression \(\hat{g} \) with target dimension \(k\).
\end{enumerate}

\begin{wrapfigure}[14]{r}{0.4\textwidth}
    \centering
    \vspace{-1\intextsep}
    \def\svgwidth{\columnwidth}
    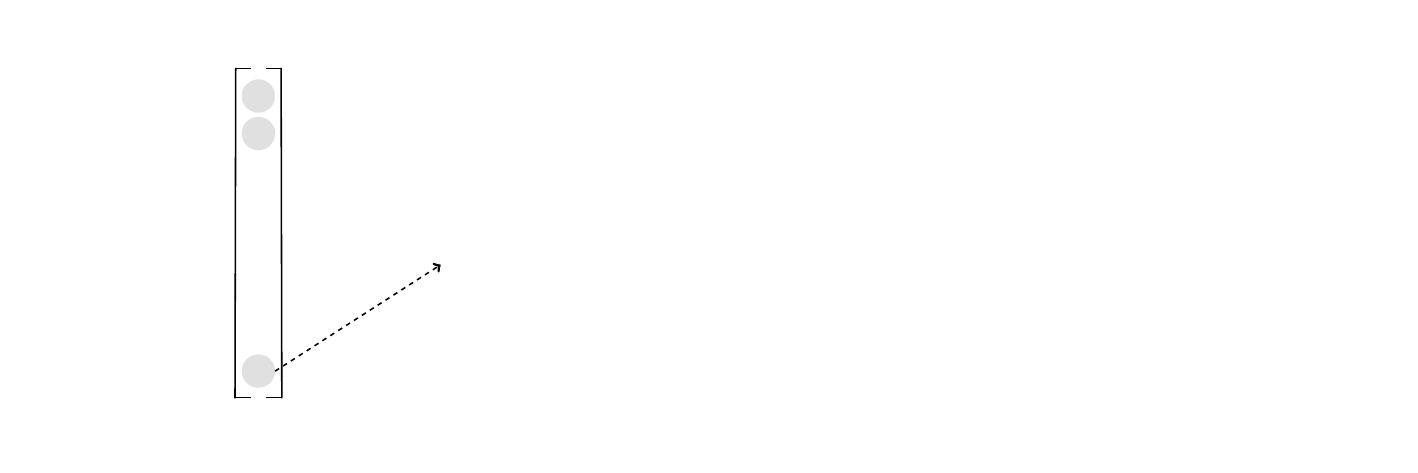

    \caption{\GraSS{}.}
    \label{fig:GraSS}
\end{wrapfigure}

We term this simple per-sample gradient compression method \emph{\textbf{Gra}dient \textbf{S}parsification and \textbf{S}parse projection} (\GraSS{}), which is illustrated in \Cref{fig:GraSS}. This leads to a \emph{sub-linear} time complexity \(O(k^{\prime} + k^{\prime} ) = O(k^{\prime} )\) to the input dimension \(p\), since the runtime of SJLT depends only on its input dimension, which is now sparsified to \(k^{\prime} \) from \(p\). In the extreme cases when \(k^{\prime} = p\), \GraSS{} reduces to vanilla SJLT; while when \(k^{\prime} = k\), \GraSS{} reduces to sparsification. Notation-wise, we write \(\SJLT_k \circ \Mask_{k^{\prime}}\).

Intuitively, \(k^{\prime} \) as a hyperparameter balances the computational complexity and attribution performance: after selecting \(k^{\prime} \) coordinates of \(g\) to form \(g^{\prime} \), subsequent application of SJLT will compress \(g^{\prime} \) down to the target dimension without losing the pair-wise distance information.

\subsubsection{\FactGraSS{}: Exploiting layer-wise gradient factorization structure}\label{subsubssec:FactGraSS:exploiting-layer-wise-gradient-factorization-structure}
In addition to FIM approximation, influence function on large-scale models often leverages the \emph{layer-wise independence} assumption by approximating FIM as a block-diagonal matrix, ignoring parameter interactions across layers. Specifically, this approach decomposes the FIM as \(\diag\{F_{\hat{\theta}_1}, \cdots, F_{\hat{\theta}_L}\}\) for an \(L\)-layer neural network, where each block \(F_{\hat{\theta}_l}\) corresponds to the \(l^{\text{th}}\) layer's parameters \(\hat{\theta}_{l}\), defined as \(F_{\hat{\theta}_l} = \mathbb{E}_{z}[\nabla _{\theta _l} \ell (z; \hat{\theta}) \nabla _{\theta _l} \ell (z; \hat{\theta}) ^{\top}]\). By writing \(g_{i, l} \coloneqq \nabla_{\theta_l} \ell(z_i; \hat{\theta})\), iFVP computation can now be done layer-wise as \(\widetilde{g}_{i, l} = F_{\hat{\theta}_l}^{-1} g_{i, l}\). Furthermore, coupling this trick with gradient compression, i.e., consider compressing \(g_{i, l}\) to \(\hat{g}_{i, l}\), which subsequently forms \(\hat{F}_{\hat{\theta}_l}\), we now compute \(\widetilde{\hat{g}}_{i, l} \coloneqq \hat{F}_{\hat{\theta}_l}^{-1} \hat{g}_{i, l}\). 

However, this renders a critical challenge for \GraSS{} to demonstrate practical speedup via a direct application of each of these layer-wise compression sub-problems, since SJLT suffers from small problem sizes (\Cref{subsec:per-sample-gradient-sparsity}). Specifically, if each layer has roughly the same number of parameters, the compression problem size is reduced from \(p \times k\) to \(p/L \times k/L\) each if the compression dimension is allocated uniformly. Moreover, recent techniques such as \LoGra{}~\citep{choe2024your} further reduce the compression problem size of each layer-wise compression via \emph{gradient factorization}, making it even more difficult to integrate \GraSS{} with these SOTA methods to achieve a further speedup.

This motivates the need for a specialized adaptation of \GraSS{} that can effectively exploit a similar gradient factorization structure. To this end, we propose \FactGraSS{}, which explicitly incorporates this factorization structure to achieve even greater efficiency in gradient compression.

\begin{figure}[t]
    \centering
    \def\svgwidth{\columnwidth}
    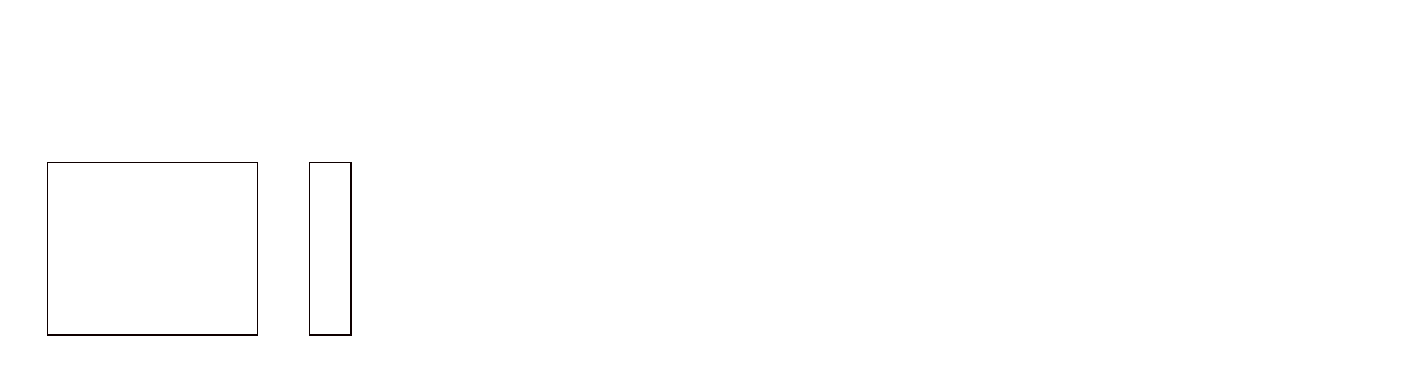

    \caption{\LoGra{} for one linear layer. Note that the application of \(\operatorname{vec}(\cdot) \) on \(\hat{g} _{i, l}\) is omitted.}
    \label{fig:LoGra}
\end{figure}

\paragraph{Recap on \LoGra{}.}
To motivate and understand the practical difficulties we must avoid, we first introduce \LoGra{}. Formally, \LoGra{} exploits the \emph{factorized structure} of linear layer's gradients. For full generality, consider a sequential input \(z_i \in \mathbb{R}^{d \times T}\) of length \(T\) to the model and the corresponding input \(z_{i,l}^{\text{in}} \in \mathbb{R}^{d_l^{\text{in}} \times T}\) and output (pre-activations) \(z_{i,l}^{\text{out}} \in \mathbb{R}^{d_l^{\text{out}} \times T}\) of the \(l^{\text{th} }\) linear layer with a weight matrix \(W_l \in \mathbb{R}^{d_l^{\text{out}} \times d_l^{\text{in}}}\) such that \(z_{i,l}^{\text{out}} = W_l z_{i,l}^{\text{in}}\). Then the gradient of the \(l^{\text{th}}\) linear layer is given by
\begin{equation}\label{eq:gradient-factorization}
    \frac{\partial \ell(z_i; \hat{\theta})}{\partial W_l}
    = \frac{\partial \ell(z_i; \hat{\theta})}{\partial z_{i,l}^{\text{out}}} \frac{\partial z_{i,l}^{\text{out}}}{\partial W_l}
    = \frac{\partial \ell(z_i; \hat{\theta})}{\partial z_{i,l}^{\text{out}}} {z_{i,l}^{\text{in}}}^{\top}
    \iff \operatorname{vec}(\mathcal{D} W_l)
    = \sum_{t=1}^{T} (z_{i,l}^{\text{in}})_{: t} \otimes \mathcal{D}(z_{i,l}^{\text{out}})_{: t},
\end{equation}
where we write \(\nabla_v \ell(z_i; \hat{\theta})\) as \(\mathcal{D} v\) for any \(v\). \LoGra{} then leverages this Kronecker-product structure of \(\operatorname{vec}(\mathcal{D} W_l)\) by assuming the projection matrix \(P_l\) for this layer has a factorized structure, i.e.,
\begin{equation}\label{eq:projected-gradient-factorization}
    P_l \operatorname{vec}(\mathcal{D} W_l)
    \coloneqq (P_l^{\text{in}} \otimes P_l^{\text{out}}) \operatorname{vec}(\mathcal{D} W_l)
    = \sum_{t=1}^{T} (P_l^{\text{in}} (z_{i,l}^{\text{in}})_{: t}) \otimes (P_l^{\text{out}} \mathcal{D} (z_{i,l}^{\text{out}})_{: t}),
\end{equation}
where \(P_l^{\text{in}} \in \mathbb{R}^{k_l^{\text{in}} \times d_l^{\text{in}}}\), \(P_l^{\text{out}} \in \mathbb{R}^{k_l^{\text{out}} \times d_l^{\text{out}}}\), and \(P_l = P_l^{\text{in}} \otimes P_l^{\text{out}} \in \mathbb{R}^{(k_l^{\text{in}} k_l^{\text{out}}) \times (d_l^{\text{in}} d_l^{\text{out}})}\). Naturally, we let \(k_l \coloneqq k_l^{\text{in} } \times k_l^{\text{out} }\) and \(p_l \coloneqq d_l^{\text{in} } \times d_l^{\text{out} }\) to be the compression dimension and number of parameters for the \(l^{\text{th} }\) linear layer, respectively. Hence, \(P_l\in \mathbb{R}^{k_l \times p_l}\) as we expect, with \(p = \sum_{l=1}^{L} p_l\) and \(k = \sum_{l=1}^{L} k_l\). The computation of \(P_l \operatorname{vec}(\mathcal{D} W_l)\) of \LoGra{} is illustrated in \Cref{fig:LoGra}, where:
\begin{itemize}[leftmargin=*]
    \item forward pass on \(z_i\) and backward pass on \(\ell(z_i; \hat{\theta})\) give \(z_{i,l}^{\text{in}}\) and \(\mathcal{D}z_{i,l}^{\text{out}}\) for each of the \(l^{\text{th} }\) linear layer;
    \item only two smaller projection problems of size \(k_l^{\text{in}} \times d_l^{\text{in}}\) and \(k_l^{\text{out}} \times d_l^{\text{out}}\) are needed for each sequential index, instead of to project the entire gradient, which is of size \((k_l^{\text{in}} k_l^{\text{out}}) \times (d_l^{\text{in}} d_l^{\text{out}}) = k_l \times p_l\);
    \item in addition, the actual gradient of the layer is \emph{never} materialized (which will require computing \Cref{eq:gradient-factorization}, \(O(T p_l)\)), only the projected gradient is materialized at the end (which takes \(O(T k_l)\)).
\end{itemize}
Notation-wise, as \(P_l^{\text{in}}\) and \(P_l^{\text{out}}\) are default to Gaussian projection~\citep{choe2024your}, we write \LoGra{} as \(\Gauss_{k_l^{\text{in}} \otimes k_l^{\text{out}}}\), where \(\otimes\) indicates that the projection is done in a factorized manner. We see that overall, assuming \(d_l^{\text{in}} \approx d_l^{\text{out}} \approx \sqrt{p_l} \), choosing \(k_l^{\text{in}} \approx k_l^{\text{out}} \approx \sqrt{k_l}\) results in a speedup from \(O(k_l p_l)\) to \(O(\sqrt{k_l p_l})\) per projection (i.e., per input and per sequential index) for the \(l^{\text{th} }\) layer.

\paragraph{Bottlenecks of integrating \GraSS{} with \LoGra{}.}
One can change the (dense) Gaussian projection used in \LoGra{} to other compression methods, e.g., SJLT, Mask, or \GraSS{}, resulting in \(\Mask_{k_l^{\text{in}} \otimes k_l^{\text{out}}}\), \(\SJLT_{k_l^{\text{in}} \otimes k_l^{\text{out}}}\), and \(\GraSS_{k_l^{\text{in}} \otimes k_l^{\text{out}}}\), respectively. However, a trivial integration with \GraSS{} will \textbf{not} lead to a practical speed up compared to \LoGra{} since each compression problem size is now reduced, making \GraSS{} slower than Gaussian projection due to the practical implementation overhead of one of its algorithmic components, SJLT, as shown in \Cref{fig:SJLT-benchmark}. This small projection problem size regime makes a direct integration of \GraSS{} with \LoGra{} challenging.

A natural idea to mitigate this issue is to apply SJLT to a \textbf{moderate} dimension by \textbf{not} factorizing the projection: by first constructing the gradients of the layer via \Cref{eq:gradient-factorization}, we can perform \GraSS{} on a much larger problem size (\(p_l \times k_l\)) rather than the two smaller problems (roughly \(\sqrt{p_l} \times \sqrt{k_l} \)). However, this results in another bottleneck: materializing the gradients explicitly blows up the space and time complexity to \(O(p_l)\), which is slower than \LoGra{}, defeating the whole purpose.

\begin{figure}[t]
    \centering
    \def\svgwidth{\columnwidth}
    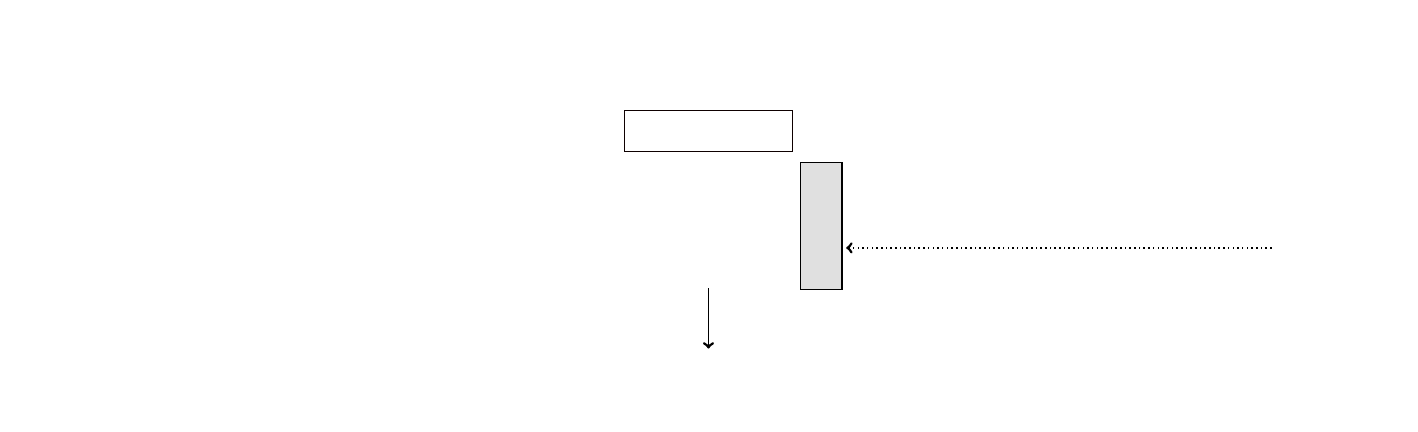

    \caption{\FactGraSS{} for one linear layer. Note that the output of SJLT is a vector in practice.}
    \label{fig:FactGraSS}
\end{figure}

\paragraph{Factorized \GraSS{}\includegraphics[height=1em]{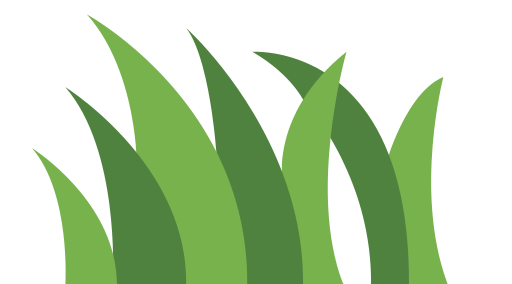}.}
To bypass the bottlenecks, we propose \emph{\textbf{Fact}orized} \GraSS{} (\FactGraSS{}), a variant of \GraSS{} that exploits the Kronecker-product structure. Specifically, given a target compression dimension \(k_l = k_l^{\text{in} } \times k_l^{\text{out} } \ll p_l\) for the \(l^{\text{th} }\) layer, after a forward pass on \(z_i\) and a backward pass on \(\ell(z_i; \hat{\theta})\) to get \(z_{i,l}^{\text{in}}\) and \(\mathcal{D}z_{i,l}^{\text{out}}\), \FactGraSS{} operates in three stages (\Cref{fig:FactGraSS}):
\begin{enumerate}[leftmargin=*]
    \item \textbf{\emph{Sparsification}}: sparsify both \(z_{i,l}^{\text{in}}\) and \(\mathcal{D} z_{i,l}^{\text{out}}\) to an intermediate dimension \({k_l^{\text{in} }}^{\prime} \) and \({k_l^{\text{out} }}^{\prime} \), where \(k_l^{\text{in} } \leq {k_l^{\text{in} }}^{\prime} \ll d_l^{\text{in} }\) and \(k_l^{\text{out} } \leq {k_l^{\text{out} }}^{\prime} \ll d_l^{\text{out} }\), respectively;
    \item \textbf{\emph{Reconstruction}}: construct the ``sparsified gradient'' \(g_{i, l}^{\prime} \) of dimension \(k_l^{\prime} \coloneqq {k_l^{\text{in} }}^{\prime} \times {k_l^{\text{out} }}^{\prime} \) via \Cref{eq:projected-gradient-factorization}, i.e., Kronecker product between the sparsified \(z_{i,l}^{\text{in}}\) and the sparsified \(\mathcal{D}z_{i,l}^{\text{out}}\);
    \item \textbf{\emph{Sparse projection}}: apply SJLT to \(g_{i, l}^{\prime} \) to get the compressed \(\hat{g} _{i, l}\) with target dimension \(k_l\).
\end{enumerate}

Intuitively, \FactGraSS{} resolves the two bottlenecks by
\begin{enumerate*}[label=\arabic*.)]
    \item avoid reconstructing the \emph{full gradient} via sparsification, and
    \item avoid small problem size for SJLT via reconstruction.
\end{enumerate*}
In terms of complexity, sparsification takes \(O({k_l^{\text{in} }}^{\prime})\) and \(O({k_l^{\text{out} }}^{\prime} )\) respectively, and reconstruction takes \(O(k_l^{\prime} )\) where \(k_l^{\prime} \coloneqq {k_l^{\text{in} }}^{\prime} \times {k_l^{\text{out} }}^{\prime}\) for performing the Kronecker product between two vectors of size \({k_l^{\text{in} }}^{\prime}\) and \({k_l^{\text{out} }}^{\prime}\), finally sparse projection also takes \(O(k_l^{\prime} )\), giving an overall time and space complexity of \(O(k_l^{\prime} )\).

Notation-wise, we write \FactGraSS{} as \(\SJLT_{k_l} \circ \Mask_{{k_l^{\text{in} }}^{\prime} \otimes {k_l^{\text{out} }}^{\prime}}\), where we use \(\otimes\) to indicate that the sparsification is done in a factorized manner. The following summarizes the complexity of our proposed methods (\GraSS{} and \FactGraSS{}), as well as other baselines we have mentioned:

\resizebox{\textwidth}{!}{
    \begin{tabular}{c cc cc c cc c cc c c c c}
        \toprule
                             &  & \multicolumn{8}{c}{\emph{\textbf{General}}} &  & \multicolumn{3}{c}{\emph{\textbf{Linear Layer}}}                                                                                                                                                                                                                                                                                                                          \\\cline{3-10}\cline{12-14}\addlinespace
                             &  & \textbf{Sparsification}                     &  & \textbf{Sparse Projection}                       &  & \textbf{\GraSS{}}                    &  & \multicolumn{2}{c}{\textbf{Baselines}} &                    & \textbf{\FactGraSS{}} &                                                                                              & \textbf{\LoGra{} (Baseline)}                                                         \\
        \textbf{Compression} &  & \(\Mask_k\)                                 &  & \(\SJLT_k\)                                      &  & \(\SJLT_k \circ \Mask_{k^{\prime}}\) &  & \(\Gauss_k\)                           & \(\FJLT_k\)        &                       & \(\SJLT_{k_l} \circ \Mask_{{k_l^{\text{in}}}^{\prime} \otimes {k_l^{\text{out}}}^{\prime}}\) &                              & \(\Gauss_{k_l^{\text{in}} \otimes k_l^{\text{out}}}\) \\
        \midrule
        Complexity           &  & \(O(k)\)                                    &  & \(O(p)\)                                         &  & \(O(k^{\prime})\)                    &  & \(O(pk)\)                              & \(O((p+k)\log p)\) &                       & \(O(k_l^{\prime})\) {\footnotesize per-layer}                                                &                              & \(O(\sqrt{p_l k_l})\) {\footnotesize per-layer}       \\
        \bottomrule
    \end{tabular}
}

In particular, compared to \LoGra{}, by writing \(k_l^{\prime} \in [k_l , p_l]\) as \(k_l^{\prime} = ck_l\) for some \emph{blow-up factor} \(c \geq 1\), \FactGraSS{} is theoretically faster than \LoGra{} if \(k_l^{\prime} \leq \sqrt{k_l p_l}\), or equivalently, \(c \leq \sqrt{p_l / k_l}\). In practice, this is easy to satisfy: for instance, consider a linear layer of size \(p_l = 4096 \times 4096\) with \(k_l = 64 \times 64\). In this case, for any blow-up factor \(c\leq 64\), \textnormal{\FactGraSS{}} is faster than \textnormal{\LoGra{}}.

With \(k^{\prime} \coloneqq \sum_{l=1}^{L} k_l^{\prime} \), \FactGraSS{} overall takes \(O(k^{\prime} )\) per compression, same as \GraSS{} \textbf{without} ever materializing the full gradient. We summarize both \GraSS{} and \FactGraSS{} in \Cref{thm:GraSS-and-FactGraSS}:

\begin{theorem}\label{thm:GraSS-and-FactGraSS}
    There is a \emph{sub-linear} compression algorithm with a complexity of \(O(k^{\prime} )\) per sample, where \(k < k^{\prime} \ll p\). Moreover, this extends to linear layers, where gradients are never materialized.
\end{theorem}

We conclude by noting that much of the recent gradient-based data attribution literature could benefit from per-sample gradient compression, underscoring the broad applicability of our methods.

%% file: Figures/matrix-multiplication.pdf_tex
\begingroup%
  \makeatletter%
  \providecommand\color[2][]{%
    \errmessage{(Inkscape) Color is used for the text in Inkscape, but the package 'color.sty' is not loaded}%
    \renewcommand\color[2][]{}%
  }%
  \providecommand\transparent[1]{%
    \errmessage{(Inkscape) Transparency is used (non-zero) for the text in Inkscape, but the package 'transparent.sty' is not loaded}%
    \renewcommand\transparent[1]{}%
  }%
  \providecommand\rotatebox[2]{#2}%
  \newcommand*\fsize{\dimexpr\f@size pt\relax}%
  \newcommand*\lineheight[1]{\fontsize{\fsize}{#1\fsize}\selectfont}%
  \ifx\svgwidth\undefined%
    \setlength{\unitlength}{680.31496063bp}%
    \ifx\svgscale\undefined%
      \relax%
    \else%
      \setlength{\unitlength}{\unitlength * \real{\svgscale}}%
    \fi%
  \else%
    \setlength{\unitlength}{\svgwidth}%
  \fi%
  \global\let\svgwidth\undefined%
  \global\let\svgscale\undefined%
  \makeatother%
  \begin{picture}(1,0.3)%
    \lineheight{1}%
    \setlength\tabcolsep{0pt}%
    \put(0.22176267,0.12032642){\color[rgb]{0,0,0}\makebox(0,0)[lt]{\lineheight{1.25}\smash{\begin{tabular}[t]{l}\(\scriptstyle k \times p\)\end{tabular}}}}%
    \put(0.35071383,0.00346762){\color[rgb]{0,0,0}\makebox(0,0)[lt]{\lineheight{1.25}\smash{\begin{tabular}[t]{l}\(\scriptstyle p\)\end{tabular}}}}%
    \put(0,0){\includegraphics[width=\unitlength,page=1]{matrix-multiplication.pdf}}%
    \put(0.11928752,0.28339265){\color[rgb]{0,0,0}\makebox(0,0)[lt]{\lineheight{1.25}\smash{\begin{tabular}[t]{l}\(\scriptstyle P\)\end{tabular}}}}%
    \put(0,0){\includegraphics[width=\unitlength,page=2]{matrix-multiplication.pdf}}%
    \put(0.31445026,0.2834479){\color[rgb]{0,0,0}\makebox(0,0)[lt]{\lineheight{1.25}\smash{\begin{tabular}[t]{l}\(\scriptstyle g\)\end{tabular}}}}%
    \put(0.11264248,0.18097744){\color[rgb]{0,0,0}\makebox(0,0)[lt]{\lineheight{1.25}\smash{\begin{tabular}[t]{l}\(\ddots\)\end{tabular}}}}%
    \put(0.31921711,0.11219788){\color[rgb]{0,0,0}\makebox(0,0)[lt]{\lineheight{1.25}\smash{\begin{tabular}[t]{l}\(\vdots\)\end{tabular}}}}%
    \put(0,0){\includegraphics[width=\unitlength,page=3]{matrix-multiplication.pdf}}%
    \put(0.36209459,0.19225146){\color[rgb]{0,0,0}\makebox(0,0)[lt]{\lineheight{1.25}\smash{\begin{tabular}[t]{l}\(=\)\end{tabular}}}}%
    \put(0.46572297,0.19225146){\color[rgb]{0,0,0}\makebox(0,0)[lt]{\lineheight{1.25}\smash{\begin{tabular}[t]{l}\(\times\)\end{tabular}}}}%
    \put(0,0){\includegraphics[width=\unitlength,page=4]{matrix-multiplication.pdf}}%
    \put(0.56273708,0.19225146){\color[rgb]{0,0,0}\makebox(0,0)[lt]{\lineheight{1.25}\smash{\begin{tabular}[t]{l}\(+\cdots+\)\end{tabular}}}}%
    \put(0.41038513,0.2834479){\color[rgb]{0,0,0}\makebox(0,0)[lt]{\lineheight{1.25}\smash{\begin{tabular}[t]{l}\(\scriptstyle P_{:1}\times g(1)\)\end{tabular}}}}%
    \put(0.73229491,0.2834479){\color[rgb]{0,0,0}\makebox(0,0)[lt]{\lineheight{1.25}\smash{\begin{tabular}[t]{l}\(\scriptstyle P_{:p}\times g(p)\)\end{tabular}}}}%
    \put(0,0){\includegraphics[width=\unitlength,page=5]{matrix-multiplication.pdf}}%
    \put(0.99232794,0.12694136){\color[rgb]{0,0,0}\makebox(0,0)[lt]{\lineheight{1.25}\smash{\begin{tabular}[t]{l}\(\scriptstyle k\)\end{tabular}}}}%
    \put(0.96047459,0.28466066){\color[rgb]{0,0,0}\makebox(0,0)[lt]{\lineheight{1.25}\smash{\begin{tabular}[t]{l}\(\scriptstyle\hat{g}\)\end{tabular}}}}%
    \put(0,0){\includegraphics[width=\unitlength,page=6]{matrix-multiplication.pdf}}%
    \put(0.46572297,0.19225146){\color[rgb]{0,0,0}\makebox(0,0)[lt]{\lineheight{1.25}\smash{\begin{tabular}[t]{l}\(\times\)\end{tabular}}}}%
    \put(0,0){\includegraphics[width=\unitlength,page=7]{matrix-multiplication.pdf}}%
    \put(0.79204241,0.19225146){\color[rgb]{0,0,0}\makebox(0,0)[lt]{\lineheight{1.25}\smash{\begin{tabular}[t]{l}\(\times\)\end{tabular}}}}%
    \put(0.89787572,0.19225146){\color[rgb]{0,0,0}\makebox(0,0)[lt]{\lineheight{1.25}\smash{\begin{tabular}[t]{l}\(=\)\end{tabular}}}}%
    \put(0.26287575,0.19225146){\color[rgb]{0,0,0}\makebox(0,0)[lt]{\lineheight{1.25}\smash{\begin{tabular}[t]{l}\(\times\)\end{tabular}}}}%
    \put(0.26508061,0.28339265){\color[rgb]{0,0,0}\makebox(0,0)[lt]{\lineheight{1.25}\smash{\begin{tabular}[t]{l}\(\scriptstyle \times\)\end{tabular}}}}%
    \put(0.36429946,0.28339265){\color[rgb]{0,0,0}\makebox(0,0)[lt]{\lineheight{1.25}\smash{\begin{tabular}[t]{l}\(\scriptstyle =\)\end{tabular}}}}%
    \put(0.90008055,0.28339265){\color[rgb]{0,0,0}\makebox(0,0)[lt]{\lineheight{1.25}\smash{\begin{tabular}[t]{l}\(\scriptstyle =\)\end{tabular}}}}%
    \put(0,0){\includegraphics[width=\unitlength,page=8]{matrix-multiplication.pdf}}%
    \put(0.59140021,0.28339265){\color[rgb]{0,0,0}\makebox(0,0)[lt]{\lineheight{1.25}\smash{\begin{tabular}[t]{l}\(\scriptstyle +\cdots+\)\end{tabular}}}}%
  \end{picture}%
\endgroup%

%% file: Figures/SJLT-matrix-multiplication.pdf_tex
\begingroup%
  \makeatletter%
  \providecommand\color[2][]{%
    \errmessage{(Inkscape) Color is used for the text in Inkscape, but the package 'color.sty' is not loaded}%
    \renewcommand\color[2][]{}%
  }%
  \providecommand\transparent[1]{%
    \errmessage{(Inkscape) Transparency is used (non-zero) for the text in Inkscape, but the package 'transparent.sty' is not loaded}%
    \renewcommand\transparent[1]{}%
  }%
  \providecommand\rotatebox[2]{#2}%
  \newcommand*\fsize{\dimexpr\f@size pt\relax}%
  \newcommand*\lineheight[1]{\fontsize{\fsize}{#1\fsize}\selectfont}%
  \ifx\svgwidth\undefined%
    \setlength{\unitlength}{680.31496063bp}%
    \ifx\svgscale\undefined%
      \relax%
    \else%
      \setlength{\unitlength}{\unitlength * \real{\svgscale}}%
    \fi%
  \else%
    \setlength{\unitlength}{\svgwidth}%
  \fi%
  \global\let\svgwidth\undefined%
  \global\let\svgscale\undefined%
  \makeatother%
  \begin{picture}(1,0.3)%
    \lineheight{1}%
    \setlength\tabcolsep{0pt}%
    \put(0.22176267,0.12253126){\color[rgb]{0,0,0}\makebox(0,0)[lt]{\lineheight{1.25}\smash{\begin{tabular}[t]{l}\(\scriptstyle k \times p\)\end{tabular}}}}%
    \put(0.35071383,0.00567245){\color[rgb]{0,0,0}\makebox(0,0)[lt]{\lineheight{1.25}\smash{\begin{tabular}[t]{l}\(\scriptstyle p\)\end{tabular}}}}%
    \put(0,0){\includegraphics[width=\unitlength,page=1]{SJLT-matrix-multiplication.pdf}}%
    \put(0.11928752,0.28559748){\color[rgb]{0,0,0}\makebox(0,0)[lt]{\lineheight{1.25}\smash{\begin{tabular}[t]{l}\(\scriptstyle P\)\end{tabular}}}}%
    \put(0,0){\includegraphics[width=\unitlength,page=2]{SJLT-matrix-multiplication.pdf}}%
    \put(0.31445026,0.28565273){\color[rgb]{0,0,0}\makebox(0,0)[lt]{\lineheight{1.25}\smash{\begin{tabular}[t]{l}\(\scriptstyle g\)\end{tabular}}}}%
    \put(0.11264248,0.18318227){\color[rgb]{0,0,0}\makebox(0,0)[lt]{\lineheight{1.25}\smash{\begin{tabular}[t]{l}\(\ddots\)\end{tabular}}}}%
    \put(0.31921711,0.11440271){\color[rgb]{0,0,0}\makebox(0,0)[lt]{\lineheight{1.25}\smash{\begin{tabular}[t]{l}\(\vdots\)\end{tabular}}}}%
    \put(0,0){\includegraphics[width=\unitlength,page=3]{SJLT-matrix-multiplication.pdf}}%
    \put(0.36209459,0.19445629){\color[rgb]{0,0,0}\makebox(0,0)[lt]{\lineheight{1.25}\smash{\begin{tabular}[t]{l}\(=\)\end{tabular}}}}%
    \put(0.46572297,0.19445629){\color[rgb]{0,0,0}\makebox(0,0)[lt]{\lineheight{1.25}\smash{\begin{tabular}[t]{l}\(\times\)\end{tabular}}}}%
    \put(0,0){\includegraphics[width=\unitlength,page=4]{SJLT-matrix-multiplication.pdf}}%
    \put(0.56273708,0.19445629){\color[rgb]{0,0,0}\makebox(0,0)[lt]{\lineheight{1.25}\smash{\begin{tabular}[t]{l}\(+\cdots+\)\end{tabular}}}}%
    \put(0.41038513,0.28565273){\color[rgb]{0,0,0}\makebox(0,0)[lt]{\lineheight{1.25}\smash{\begin{tabular}[t]{l}\(\scriptstyle P_{:1}\times g(1)\)\end{tabular}}}}%
    \put(0.73229491,0.28565273){\color[rgb]{0,0,0}\makebox(0,0)[lt]{\lineheight{1.25}\smash{\begin{tabular}[t]{l}\(\scriptstyle P_{:p}\times g(p)\)\end{tabular}}}}%
    \put(0,0){\includegraphics[width=\unitlength,page=5]{SJLT-matrix-multiplication.pdf}}%
    \put(0.99232794,0.12914619){\color[rgb]{0,0,0}\makebox(0,0)[lt]{\lineheight{1.25}\smash{\begin{tabular}[t]{l}\(\scriptstyle k\)\end{tabular}}}}%
    \put(0.96047459,0.28686549){\color[rgb]{0,0,0}\makebox(0,0)[lt]{\lineheight{1.25}\smash{\begin{tabular}[t]{l}\(\scriptstyle\hat{g}\)\end{tabular}}}}%
    \put(0,0){\includegraphics[width=\unitlength,page=6]{SJLT-matrix-multiplication.pdf}}%
    \put(0.46572297,0.19445629){\color[rgb]{0,0,0}\makebox(0,0)[lt]{\lineheight{1.25}\smash{\begin{tabular}[t]{l}\(\times\)\end{tabular}}}}%
    \put(0,0){\includegraphics[width=\unitlength,page=7]{SJLT-matrix-multiplication.pdf}}%
    \put(0.79204241,0.19445629){\color[rgb]{0,0,0}\makebox(0,0)[lt]{\lineheight{1.25}\smash{\begin{tabular}[t]{l}\(\times\)\end{tabular}}}}%
    \put(0.89787572,0.19445629){\color[rgb]{0,0,0}\makebox(0,0)[lt]{\lineheight{1.25}\smash{\begin{tabular}[t]{l}\(=\)\end{tabular}}}}%
    \put(0,0){\includegraphics[width=\unitlength,page=8]{SJLT-matrix-multiplication.pdf}}%
    \put(0.84174493,0.01562227){\color[rgb]{0,0,0}\makebox(0,0)[lt]{\lineheight{1.25}\smash{\begin{tabular}[t]{l}\(\scriptstyle 1\)\end{tabular}}}}%
    \put(0.93214461,0.01562227){\color[rgb]{0,0,0}\makebox(0,0)[lt]{\lineheight{1.25}\smash{\begin{tabular}[t]{l}\(\scriptstyle -1\)\end{tabular}}}}%
    \put(0,0){\includegraphics[width=\unitlength,page=9]{SJLT-matrix-multiplication.pdf}}%
    \put(0.26287575,0.19445629){\color[rgb]{0,0,0}\makebox(0,0)[lt]{\lineheight{1.25}\smash{\begin{tabular}[t]{l}\(\times\)\end{tabular}}}}%
    \put(0.26508061,0.28559748){\color[rgb]{0,0,0}\makebox(0,0)[lt]{\lineheight{1.25}\smash{\begin{tabular}[t]{l}\(\scriptstyle \times\)\end{tabular}}}}%
    \put(0.36429946,0.28559748){\color[rgb]{0,0,0}\makebox(0,0)[lt]{\lineheight{1.25}\smash{\begin{tabular}[t]{l}\(\scriptstyle =\)\end{tabular}}}}%
    \put(0.90008055,0.28559748){\color[rgb]{0,0,0}\makebox(0,0)[lt]{\lineheight{1.25}\smash{\begin{tabular}[t]{l}\(\scriptstyle =\)\end{tabular}}}}%
    \put(0.59140021,0.28559748){\color[rgb]{0,0,0}\makebox(0,0)[lt]{\lineheight{1.25}\smash{\begin{tabular}[t]{l}\(\scriptstyle +\cdots+\)\end{tabular}}}}%
  \end{picture}%
\endgroup%

%% file: Figures/SJLT.pdf_tex
\begingroup%
  \makeatletter%
  \providecommand\color[2][]{%
    \errmessage{(Inkscape) Color is used for the text in Inkscape, but the package 'color.sty' is not loaded}%
    \renewcommand\color[2][]{}%
  }%
  \providecommand\transparent[1]{%
    \errmessage{(Inkscape) Transparency is used (non-zero) for the text in Inkscape, but the package 'transparent.sty' is not loaded}%
    \renewcommand\transparent[1]{}%
  }%
  \providecommand\rotatebox[2]{#2}%
  \newcommand*\fsize{\dimexpr\f@size pt\relax}%
  \newcommand*\lineheight[1]{\fontsize{\fsize}{#1\fsize}\selectfont}%
  \ifx\svgwidth\undefined%
    \setlength{\unitlength}{680.31496063bp}%
    \ifx\svgscale\undefined%
      \relax%
    \else%
      \setlength{\unitlength}{\unitlength * \real{\svgscale}}%
    \fi%
  \else%
    \setlength{\unitlength}{\svgwidth}%
  \fi%
  \global\let\svgwidth\undefined%
  \global\let\svgscale\undefined%
  \makeatother%
  \begin{picture}(1,0.28333333)%
    \lineheight{1}%
    \setlength\tabcolsep{0pt}%
    \put(0.00169762,0.13899638){\color[rgb]{0,0,0}\makebox(0,0)[lt]{\lineheight{1.25}\smash{\begin{tabular}[t]{l}\(g\)\end{tabular}}}}%
    \put(0.24202749,0.13906898){\color[rgb]{0,0,0}\makebox(0,0)[lt]{\lineheight{1.25}\smash{\begin{tabular}[t]{l}\(\hat{g}\)\end{tabular}}}}%
    \put(0,0){\includegraphics[width=\unitlength,page=1]{SJLT.pdf}}%
    \put(0.2308095,0.25975386){\color[rgb]{0,0,0}\makebox(0,0)[lt]{\lineheight{1.25}\smash{\begin{tabular}[t]{l}\(\times 1\)\end{tabular}}}}%
    \put(0,0){\includegraphics[width=\unitlength,page=2]{SJLT.pdf}}%
    \put(0.2308095,0.23329563){\color[rgb]{0,0,0}\makebox(0,0)[lt]{\lineheight{1.25}\smash{\begin{tabular}[t]{l}\(\times -1\)\end{tabular}}}}%
    \put(0.23431363,0.0699536){\color[rgb]{0,0,0}\makebox(0,0)[lt]{\lineheight{1.25}\smash{\begin{tabular}[t]{l}\(k\)\end{tabular}}}}%
    \put(0,0){\includegraphics[width=\unitlength,page=3]{SJLT.pdf}}%
    \put(0.07412548,0.00218786){\color[rgb]{0,0,0}\makebox(0,0)[lt]{\lineheight{1.25}\smash{\begin{tabular}[t]{l}\(p\)\end{tabular}}}}%
    \put(0,0){\includegraphics[width=\unitlength,page=4]{SJLT.pdf}}%
    \put(0.04145442,0.12969348){\color[rgb]{0,0,0}\makebox(0,0)[lt]{\lineheight{1.25}\smash{\begin{tabular}[t]{l}\(\vdots\)\end{tabular}}}}%
    \put(0,0){\includegraphics[width=\unitlength,page=5]{SJLT.pdf}}%
  \end{picture}%
\endgroup%

%% file: Figures/Mask.pdf_tex
\begingroup%
  \makeatletter%
  \providecommand\color[2][]{%
    \errmessage{(Inkscape) Color is used for the text in Inkscape, but the package 'color.sty' is not loaded}%
    \renewcommand\color[2][]{}%
  }%
  \providecommand\transparent[1]{%
    \errmessage{(Inkscape) Transparency is used (non-zero) for the text in Inkscape, but the package 'transparent.sty' is not loaded}%
    \renewcommand\transparent[1]{}%
  }%
  \providecommand\rotatebox[2]{#2}%
  \newcommand*\fsize{\dimexpr\f@size pt\relax}%
  \newcommand*\lineheight[1]{\fontsize{\fsize}{#1\fsize}\selectfont}%
  \ifx\svgwidth\undefined%
    \setlength{\unitlength}{680.31496063bp}%
    \ifx\svgscale\undefined%
      \relax%
    \else%
      \setlength{\unitlength}{\unitlength * \real{\svgscale}}%
    \fi%
  \else%
    \setlength{\unitlength}{\svgwidth}%
  \fi%
  \global\let\svgwidth\undefined%
  \global\let\svgscale\undefined%
  \makeatother%
  \begin{picture}(1,0.275)%
    \lineheight{1}%
    \setlength\tabcolsep{0pt}%
    \put(0.00169762,0.13507271){\color[rgb]{0,0,0}\makebox(0,0)[lt]{\lineheight{1.25}\smash{\begin{tabular}[t]{l}\(g\)\end{tabular}}}}%
    \put(0.24202749,0.13514531){\color[rgb]{0,0,0}\makebox(0,0)[lt]{\lineheight{1.25}\smash{\begin{tabular}[t]{l}\(\hat{g}\)\end{tabular}}}}%
    \put(0,0){\includegraphics[width=\unitlength,page=1]{Mask.pdf}}%
    \put(0.2308095,0.25583019){\color[rgb]{0,0,0}\makebox(0,0)[lt]{\lineheight{1.25}\smash{\begin{tabular}[t]{l}\(\times 1\)\end{tabular}}}}%
    \put(0.24092822,0.05941544){\color[rgb]{0,0,0}\makebox(0,0)[lt]{\lineheight{1.25}\smash{\begin{tabular}[t]{l}\(k\)\end{tabular}}}}%
    \put(0,0){\includegraphics[width=\unitlength,page=2]{Mask.pdf}}%
    \put(0.06971575,0.00267385){\color[rgb]{0,0,0}\makebox(0,0)[lt]{\lineheight{1.25}\smash{\begin{tabular}[t]{l}\(p\)\end{tabular}}}}%
    \put(0,0){\includegraphics[width=\unitlength,page=3]{Mask.pdf}}%
    \put(0.04145442,0.12576981){\color[rgb]{0,0,0}\makebox(0,0)[lt]{\lineheight{1.25}\smash{\begin{tabular}[t]{l}\(\vdots\)\end{tabular}}}}%
    \put(0,0){\includegraphics[width=\unitlength,page=4]{Mask.pdf}}%
  \end{picture}%
\endgroup%

%% file: Figures/GraSS.pdf_tex
\begingroup%
  \makeatletter%
  \providecommand\color[2][]{%
    \errmessage{(Inkscape) Color is used for the text in Inkscape, but the package 'color.sty' is not loaded}%
    \renewcommand\color[2][]{}%
  }%
  \providecommand\transparent[1]{%
    \errmessage{(Inkscape) Transparency is used (non-zero) for the text in Inkscape, but the package 'transparent.sty' is not loaded}%
    \renewcommand\transparent[1]{}%
  }%
  \providecommand\rotatebox[2]{#2}%
  \newcommand*\fsize{\dimexpr\f@size pt\relax}%
  \newcommand*\lineheight[1]{\fontsize{\fsize}{#1\fsize}\selectfont}%
  \ifx\svgwidth\undefined%
    \setlength{\unitlength}{680.31496063bp}%
    \ifx\svgscale\undefined%
      \relax%
    \else%
      \setlength{\unitlength}{\unitlength * \real{\svgscale}}%
    \fi%
  \else%
    \setlength{\unitlength}{\svgwidth}%
  \fi%
  \global\let\svgwidth\undefined%
  \global\let\svgscale\undefined%
  \makeatother%
  \begin{picture}(1,0.33333333)%
    \lineheight{1}%
    \setlength\tabcolsep{0pt}%
    \put(0,0){\includegraphics[width=\unitlength,page=1]{GraSS.pdf}}%
    \put(0.07103597,0.00396271){\color[rgb]{0,0,0}\makebox(0,0)[lt]{\lineheight{1.25}\smash{\begin{tabular}[t]{l}\(p\)\end{tabular}}}}%
    \put(0.21316829,0.04173965){\color[rgb]{0,0,0}\makebox(0,0)[lt]{\lineheight{1.25}\smash{\begin{tabular}[t]{l}\(k'\)\end{tabular}}}}%
    \put(0,0){\includegraphics[width=\unitlength,page=2]{GraSS.pdf}}%
    \put(0.33688326,0.31728942){\color[rgb]{0,0,0}\makebox(0,0)[lt]{\lineheight{1.25}\smash{\begin{tabular}[t]{l}\(\times 1\)\end{tabular}}}}%
    \put(0,0){\includegraphics[width=\unitlength,page=3]{GraSS.pdf}}%
    \put(0.33688326,0.29524078){\color[rgb]{0,0,0}\makebox(0,0)[lt]{\lineheight{1.25}\smash{\begin{tabular}[t]{l}\(\times -1\)\end{tabular}}}}%
    \put(0.04248809,0.11640294){\color[rgb]{0,0,0}\makebox(0,0)[lt]{\lineheight{1.25}\smash{\begin{tabular}[t]{l}\(\vdots\)\end{tabular}}}}%
    \put(-0.00026648,0.16319911){\color[rgb]{0,0,0}\makebox(0,0)[lt]{\lineheight{1.25}\smash{\begin{tabular}[t]{l}\(g\)\end{tabular}}}}%
    \put(0,0){\includegraphics[width=\unitlength,page=4]{GraSS.pdf}}%
    \put(0.33995082,0.09219138){\color[rgb]{0,0,0}\makebox(0,0)[lt]{\lineheight{1.25}\smash{\begin{tabular}[t]{l}\(k\)\end{tabular}}}}%
    \put(0.34766483,0.16317978){\color[rgb]{0,0,0}\makebox(0,0)[lt]{\lineheight{1.25}\smash{\begin{tabular}[t]{l}\(\hat{g}\)\end{tabular}}}}%
    \put(0,0){\includegraphics[width=\unitlength,page=5]{GraSS.pdf}}%
    \put(0.07827842,0.31713562){\color[rgb]{0,0,0}\makebox(0,0)[lt]{\lineheight{1.25}\smash{\begin{tabular}[t]{l}\(\Mask_{k'}\)\end{tabular}}}}%
    \put(0.2255325,0.26356023){\color[rgb]{0,0,0}\makebox(0,0)[lt]{\lineheight{1.25}\smash{\begin{tabular}[t]{l}\(\SJLT_k\)\end{tabular}}}}%
    \put(0,0){\includegraphics[width=\unitlength,page=6]{GraSS.pdf}}%
    \put(0.17882201,0.15168088){\color[rgb]{0,0,0}\makebox(0,0)[lt]{\lineheight{1.25}\smash{\begin{tabular}[t]{l}\(\vdots\)\end{tabular}}}}%
    \put(0.13202519,0.16319911){\color[rgb]{0,0,0}\makebox(0,0)[lt]{\lineheight{1.25}\smash{\begin{tabular}[t]{l}\(g'\)\end{tabular}}}}%
  \end{picture}%
\endgroup%

%% file: Figures/LoGra.pdf_tex
\begingroup%
  \makeatletter%
  \providecommand\color[2][]{%
    \errmessage{(Inkscape) Color is used for the text in Inkscape, but the package 'color.sty' is not loaded}%
    \renewcommand\color[2][]{}%
  }%
  \providecommand\transparent[1]{%
    \errmessage{(Inkscape) Transparency is used (non-zero) for the text in Inkscape, but the package 'transparent.sty' is not loaded}%
    \renewcommand\transparent[1]{}%
  }%
  \providecommand\rotatebox[2]{#2}%
  \newcommand*\fsize{\dimexpr\f@size pt\relax}%
  \newcommand*\lineheight[1]{\fontsize{\fsize}{#1\fsize}\selectfont}%
  \ifx\svgwidth\undefined%
    \setlength{\unitlength}{680.31496063bp}%
    \ifx\svgscale\undefined%
      \relax%
    \else%
      \setlength{\unitlength}{\unitlength * \real{\svgscale}}%
    \fi%
  \else%
    \setlength{\unitlength}{\svgwidth}%
  \fi%
  \global\let\svgwidth\undefined%
  \global\let\svgscale\undefined%
  \makeatother%
  \begin{picture}(1,0.27083333)%
    \lineheight{1}%
    \setlength\tabcolsep{0pt}%
    \put(0,0){\includegraphics[width=\unitlength,page=1]{LoGra.pdf}}%
    \put(0.0916317,0.08962262){\color[rgb]{0,0,0}\makebox(0,0)[lt]{\lineheight{1.25}\smash{\begin{tabular}[t]{l}\(W_l\)\end{tabular}}}}%
    \put(0.19009425,0.08962262){\color[rgb]{0,0,0}\makebox(0,0)[lt]{\lineheight{1.25}\smash{\begin{tabular}[t]{l}\(=\)\end{tabular}}}}%
    \put(0.03868816,0.25299187){\color[rgb]{0,0,0}\makebox(0,0)[lt]{\lineheight{1.25}\smash{\begin{tabular}[t]{l}Forward Pass\end{tabular}}}}%
    \put(0.73025036,0.25299187){\color[rgb]{0,0,0}\makebox(0,0)[lt]{\lineheight{1.25}\smash{\begin{tabular}[t]{l}Backward Pass\end{tabular}}}}%
    \put(0,0){\includegraphics[width=\unitlength,page=2]{LoGra.pdf}}%
    \put(0.46950439,0.06345978){\color[rgb]{0,0,0}\makebox(0,0)[lt]{\lineheight{1.25}\smash{\begin{tabular}[t]{l}\(\hat{g}_{i,l}\)\end{tabular}}}}%
    \put(0,0){\includegraphics[width=\unitlength,page=3]{LoGra.pdf}}%
    \put(0.33070154,0.18832239){\color[rgb]{0,0,0}\makebox(0,0)[lt]{\lineheight{1.25}\smash{\begin{tabular}[t]{l}\(P_l^{\text{in}}\)\end{tabular}}}}%
    \put(0.62588367,0.08086332){\color[rgb]{0,0,0}\makebox(0,0)[lt]{\lineheight{1.25}\smash{\begin{tabular}[t]{l}\(P_l^{\text{out}}\)\end{tabular}}}}%
    \put(0.10040149,0.20189381){\color[rgb]{0,0,0}\makebox(0,0)[lt]{\lineheight{1.25}\smash{\begin{tabular}[t]{l}\(d_l^{\text{in}}\)\end{tabular}}}}%
    \put(0.5671725,0.01838404){\color[rgb]{0,0,0}\makebox(0,0)[lt]{\lineheight{1.25}\smash{\begin{tabular}[t]{l}\(k_l^{\text{out}}\)\end{tabular}}}}%
    \put(0.49584055,0.15067984){\color[rgb]{0,0,0}\makebox(0,0)[lt]{\lineheight{1.25}\smash{\begin{tabular}[t]{l}\(k_l^{\text{in}}\)\end{tabular}}}}%
    \put(0,0){\includegraphics[width=\unitlength,page=4]{LoGra.pdf}}%
    \put(0.54189739,0.12412879){\color[rgb]{0,0,0}\makebox(0,0)[lt]{\lineheight{1.25}\smash{\begin{tabular}[t]{l}\(\otimes\)\end{tabular}}}}%
    \put(0,0){\includegraphics[width=\unitlength,page=5]{LoGra.pdf}}%
    \put(-0.00472514,0.17161013){\color[rgb]{0,0,0}\makebox(0,0)[lt]{\lineheight{1.25}\smash{\begin{tabular}[t]{l}\(z_{i,l}^{\text{in}}\)\end{tabular}}}}%
    \put(0,0){\includegraphics[width=\unitlength,page=6]{LoGra.pdf}}%
    \put(0.87300243,0.16622321){\color[rgb]{0,0,0}\makebox(0,0)[lt]{\lineheight{1.25}\smash{\begin{tabular}[t]{l}\(\mathcal{D} z_{i,l}^{\text{out} }\)\end{tabular}}}}%
    \put(0.78326486,0.08953044){\color[rgb]{0,0,0}\makebox(0,0)[lt]{\lineheight{1.25}\smash{\begin{tabular}[t]{l}\(W_l\)\end{tabular}}}}%
    \put(0.91543795,0.11832256){\color[rgb]{0,0,0}\makebox(0,0)[lt]{\lineheight{1.25}\smash{\begin{tabular}[t]{l}\(d_l^{\text{out}}\)\end{tabular}}}}%
    \put(0.79354852,0.18115349){\color[rgb]{0,0,0}\rotatebox{-90}{\makebox(0,0)[lt]{\lineheight{1.25}\smash{\begin{tabular}[t]{l}\(=\)\end{tabular}}}}}%
    \put(0.66631921,0.19351641){\color[rgb]{0,0,0}\makebox(0,0)[lt]{\lineheight{1.25}\smash{\begin{tabular}[t]{l}\(\mathcal{D}z_{i,l}^{\text{in}}\)\end{tabular}}}}%
    \put(0,0){\includegraphics[width=\unitlength,page=7]{LoGra.pdf}}%
    \put(0.21777864,0.16622321){\color[rgb]{0,0,0}\makebox(0,0)[lt]{\lineheight{1.25}\smash{\begin{tabular}[t]{l}\(z_{i,l}^{\text{out}}\)\end{tabular}}}}%
  \end{picture}%
\endgroup%

%% file: Figures/FactGraSS.pdf_tex
\begingroup%
  \makeatletter%
  \providecommand\color[2][]{%
    \errmessage{(Inkscape) Color is used for the text in Inkscape, but the package 'color.sty' is not loaded}%
    \renewcommand\color[2][]{}%
  }%
  \providecommand\transparent[1]{%
    \errmessage{(Inkscape) Transparency is used (non-zero) for the text in Inkscape, but the package 'transparent.sty' is not loaded}%
    \renewcommand\transparent[1]{}%
  }%
  \providecommand\rotatebox[2]{#2}%
  \newcommand*\fsize{\dimexpr\f@size pt\relax}%
  \newcommand*\lineheight[1]{\fontsize{\fsize}{#1\fsize}\selectfont}%
  \ifx\svgwidth\undefined%
    \setlength{\unitlength}{680.31496063bp}%
    \ifx\svgscale\undefined%
      \relax%
    \else%
      \setlength{\unitlength}{\unitlength * \real{\svgscale}}%
    \fi%
  \else%
    \setlength{\unitlength}{\svgwidth}%
  \fi%
  \global\let\svgwidth\undefined%
  \global\let\svgscale\undefined%
  \makeatother%
  \begin{picture}(1,0.30416667)%
    \lineheight{1}%
    \setlength\tabcolsep{0pt}%
    \put(0.74750137,0.28742917){\color[rgb]{0,0,0}\makebox(0,0)[lt]{\lineheight{1.25}\smash{\begin{tabular}[t]{l}Backward Pass\end{tabular}}}}%
    \put(0,0){\includegraphics[width=\unitlength,page=1]{FactGraSS.pdf}}%
    \put(0.50835381,0.068905){\color[rgb]{0,0,0}\makebox(0,0)[lt]{\lineheight{1.25}\smash{\begin{tabular}[t]{l}\(\SJLT_{k_l}\)\end{tabular}}}}%
    \put(0,0){\includegraphics[width=\unitlength,page=2]{FactGraSS.pdf}}%
    \put(0.48755824,0.02191747){\color[rgb]{0,0,0}\makebox(0,0)[lt]{\lineheight{1.25}\smash{\begin{tabular}[t]{l}\(\hat{g}_{i,l}\)\end{tabular}}}}%
    \put(0.32190558,0.24943975){\color[rgb]{0,0,0}\makebox(0,0)[lt]{\lineheight{1.25}\smash{\begin{tabular}[t]{l}\(\Mask_{{k_l^{\text{in}}}'}\)\end{tabular}}}}%
    \put(0.62988434,0.14423358){\color[rgb]{0,0,0}\makebox(0,0)[lt]{\lineheight{1.25}\smash{\begin{tabular}[t]{l}\(\Mask_{{k_l^{\text{out}}}'}\)\end{tabular}}}}%
    \put(0,0){\includegraphics[width=\unitlength,page=3]{FactGraSS.pdf}}%
    \put(0.56997725,0.20604757){\color[rgb]{0,0,0}\makebox(0,0)[lt]{\lineheight{1.25}\smash{\begin{tabular}[t]{l}\(\otimes\)\end{tabular}}}}%
    \put(0.51824853,0.23762442){\color[rgb]{0,0,0}\makebox(0,0)[lt]{\lineheight{1.25}\smash{\begin{tabular}[t]{l}\({k_l^{\text{in}}}'\)\end{tabular}}}}%
    \put(0.60236588,0.09726448){\color[rgb]{0,0,0}\makebox(0,0)[lt]{\lineheight{1.25}\smash{\begin{tabular}[t]{l}\({k_l^{\text{out}}}'\)\end{tabular}}}}%
    \put(0.53672441,0.00232468){\color[rgb]{0,0,0}\makebox(0,0)[lt]{\lineheight{1.25}\smash{\begin{tabular}[t]{l}\(k_l^{\text{in}} \times k_l^{\text{out}}\)\end{tabular}}}}%
    \put(0,0){\includegraphics[width=\unitlength,page=4]{FactGraSS.pdf}}%
    \put(0.10006318,0.14610844){\color[rgb]{0,0,0}\makebox(0,0)[lt]{\lineheight{1.25}\smash{\begin{tabular}[t]{l}\(W_l\)\end{tabular}}}}%
    \put(0.19852573,0.14610844){\color[rgb]{0,0,0}\makebox(0,0)[lt]{\lineheight{1.25}\smash{\begin{tabular}[t]{l}\(=\)\end{tabular}}}}%
    \put(0.04707077,0.28742917){\color[rgb]{0,0,0}\makebox(0,0)[lt]{\lineheight{1.25}\smash{\begin{tabular}[t]{l}Forward Pass\end{tabular}}}}%
    \put(0,0){\includegraphics[width=\unitlength,page=5]{FactGraSS.pdf}}%
    \put(0.10883297,0.25837968){\color[rgb]{0,0,0}\makebox(0,0)[lt]{\lineheight{1.25}\smash{\begin{tabular}[t]{l}\(d_l^{\text{in}}\)\end{tabular}}}}%
    \put(0,0){\includegraphics[width=\unitlength,page=6]{FactGraSS.pdf}}%
    \put(0.00370635,0.22809601){\color[rgb]{0,0,0}\makebox(0,0)[lt]{\lineheight{1.25}\smash{\begin{tabular}[t]{l}\(z_{i,l}^{\text{in}}\)\end{tabular}}}}%
    \put(0,0){\includegraphics[width=\unitlength,page=7]{FactGraSS.pdf}}%
    \put(0.89025332,0.22270915){\color[rgb]{0,0,0}\makebox(0,0)[lt]{\lineheight{1.25}\smash{\begin{tabular}[t]{l}\(\mathcal{D} z_{i,l}^{\text{out} }\)\end{tabular}}}}%
    \put(0.80051575,0.14601638){\color[rgb]{0,0,0}\makebox(0,0)[lt]{\lineheight{1.25}\smash{\begin{tabular}[t]{l}\(W_l\)\end{tabular}}}}%
    \put(0.93268878,0.1748085){\color[rgb]{0,0,0}\makebox(0,0)[lt]{\lineheight{1.25}\smash{\begin{tabular}[t]{l}\(d_l^{\text{out}}\)\end{tabular}}}}%
    \put(0.81079941,0.23763943){\color[rgb]{0,0,0}\rotatebox{-90}{\makebox(0,0)[lt]{\lineheight{1.25}\smash{\begin{tabular}[t]{l}\(=\)\end{tabular}}}}}%
    \put(0.6835701,0.25000235){\color[rgb]{0,0,0}\makebox(0,0)[lt]{\lineheight{1.25}\smash{\begin{tabular}[t]{l}\(\mathcal{D}z_{i,l}^{\text{in}}\)\end{tabular}}}}%
    \put(0.48755824,0.14005477){\color[rgb]{0,0,0}\makebox(0,0)[lt]{\lineheight{1.25}\smash{\begin{tabular}[t]{l}\(g_{i,l}^{\prime}\)\end{tabular}}}}%
    \put(0,0){\includegraphics[width=\unitlength,page=8]{FactGraSS.pdf}}%
    \put(0.22621012,0.22270908){\color[rgb]{0,0,0}\makebox(0,0)[lt]{\lineheight{1.25}\smash{\begin{tabular}[t]{l}\(z_{i,l}^{\text{out}}\)\end{tabular}}}}%
  \end{picture}%
\endgroup%

%% file: 4_experiment.tex
\section{Experiment}\label{sec:experiment}
In this section, we evaluate the effectiveness of \GraSS{} and \FactGraSS{} in terms of \emph{accuracy} and \emph{efficiency}. Specifically, in \Cref{subsec:quantitative-accuracy-with-counterfactual-evaluation}, we first perform the standard counterfactual evaluations to quantitatively study the data valuation accuracy of \GraSS{} and \FactGraSS{} on small-scale setups. Then, we scale \FactGraSS{} to a billion-scale model and billion-token dataset, where we investigate the qualitative accuracy and memory/compute efficiency in \Cref{subsec:scaling-up-to-billion-size-model}. Further experimental details, such as hyperparameters and compute resources, can be found in \Cref{adxsubsec:details-of-models-datasets-and-computing-resources}.

\subsection{Quantitative accuracy via counterfactual evaluation}\label{subsec:quantitative-accuracy-with-counterfactual-evaluation}
We assess the quantitative accuracy of data attribution algorithms using the widely adopted \emph{linear datamodeling score} (LDS)~\citep{park2023trak}, a counterfactual evaluation method. While LDS relies on the additivity assumption, which is known to be imperfect~\citep{hu2024most}, it remains a valuable evaluation metric for data attribution. All the quantitative experiments are conducted on one \texttt{NVIDIA A40 GPU} with \texttt{48 GB} memory, and other details can be found in \Cref{adxsubsec:details-of-quantitative-analysis}.

\paragraph{\GraSS{} with \Trak{}.}
We apply \GraSS{} on one of the SOTA data attribution algorithms (in terms of attribution quality), \Trak{}~\citep{park2023trak}, with the implementation from the \texttt{dattri} library~\citep{deng2024dattri}. To validate the effectiveness of our sparsification and sparse projection methods, we conduct an ablation study on a simple 3-layer MLP trained on MNIST~\citep{lecun1998mnist}. As shown in \Cref{subtab:MLP_MNIST}, even a standalone Random Mask achieves non-trivial LDS results, while Selective Mask improves the performance further. Additionally, the sparse projection SJLT significantly outperforms baselines like FJLT and Gaussian projection in both efficiency and LDS accuracy.

We evaluate \GraSS{} on more complex models:
\begin{enumerate*}[label=\arabic*.)]
    \item ResNet9~\citep{he2016deep} with CIFAR2~\citep{krizhevsky2009learning}, and
    \item Music Transformer~\citep{huang2018music} with MAESTRO~\citep{hawthorne2018enabling}.
\end{enumerate*}
Results in \Cref{subtab:ResNet9_CIFAR2,subtab:MusicTransformer_MAESTRO} demonstrate that while sparsification methods are highly efficient, they often fall short in LDS performance. In contrast, sparse projection methods achieve competitive LDS scores but typically incur higher projection costs, though they still outperform the baseline\footnote{We omit \(\Gauss_k\) since the projection matrices for these two models are too large to fit in the GPU memory.} FJLT by a large margin. Notably, \GraSS{} strikes a balance between these extremes, achieving competitive LDS scores at a fraction of the computational cost.

\paragraph{\FactGraSS{} with Influence Function on Linear Layers.}
We next evaluate \FactGraSS{} with layer-wise block-diagonal FIM influence functions for linear layers. We consider a small language model, GPT2-small~\citep{radford2019language} fine-tuned on the WikiText dataset~\citep{merity2016pointer}, to enable LDS evaluation. The results are presented in \Cref{subtab:GPT2_WikiText}, where \(k_l\) indicates the target compression dimension for each linear layer. We further set \(k_l^{\text{in} } = k_l^{\text{out} } = \sqrt{k_l} \) for simplicity. As discussed in \Cref{subsubssec:FactGraSS:exploiting-layer-wise-gradient-factorization-structure}, replacing the Gaussian projection matrices in \LoGra{} with SJLT most likely will result in an efficiency degradation, although it achieves a competitive LDS. On the other hand, standalone sparsification achieves competitive LDS results with minimal compression overhead, highlighting its potential as an efficient alternative in overparametrized models. Finally, \FactGraSS{} not only maintains the LDS performance of SJLT but also significantly improves computational efficiency, achieving up to a \(250\%\) speedup over the most efficient SOTA baseline \LoGra{}.

\begin{table}[H]
    \centering
    \caption{Quantitative evaluation results with different gradient compression methods.}
    \label{tab:quantitative-LDS}
    \begin{subtable}[c]{\textwidth}
        \centering
        \caption{LDS and compression wall-time for MLP with MNIST on \Trak{}.}
        \label{subtab:MLP_MNIST}
        \resizebox{\textwidth}{!}{
            \begin{tabular}{c ccc c ccc c ccc c ccc c ccc}
                \toprule
                         & \multicolumn{7}{c}{\textbf{Sparsification}} &            & \multicolumn{3}{c}{\textbf{Sparse Projection}} &  & \multicolumn{7}{c}{\textbf{Baselines}}                                                                                                                                                                                                                           \\\addlinespace\addlinespace[0.2em]
                         & \multicolumn{3}{c}{\(\RM_{k}\)}             &            & \multicolumn{3}{c}{\(\SM_{k}\)}                &  & \multicolumn{3}{c}{\(\SJLT_{k}\)}      &                     & \multicolumn{3}{c}{\(\FJLT_{k}\)} &  & \multicolumn{3}{c}{\(\Gauss_{k}\)}                                                                                                                         \\\cline{2-4} \cline{6-8} \cline{10-12} \cline{14-16} \cline{18-20}\addlinespace[0.2em]
                \(k\)    & \(2048\)                                    & \(4096\)   & \(8192\)                                       &  & \(2048\)                               & \(4096\)            & \(8192\)                          &  & \(2048\)                           & \(4096\)            & \(8192\)   &  & \(2048\)   & \(4096\)   & \(8192\)   &  & \(2048\)   & \(4096\)   & \(8192\)    \\
                \midrule
                LDS      & \(0.3803\)                                  & \(0.4054\) & \(0.4318\)                                     &  & \(0.3882\)                             & \(0.4163\)          & \(\mathbf{0.4373}\)               &  & \(\mathbf{0.4171}\)                & \(\mathbf{0.4280}\) & \(0.4357\) &  & \(0.4146\) & \(0.4359\) & \(0.4347\) &  & \(0.4101\) & \(0.4253\) & \(0.4346\)  \\
                Time (s) & \(0.1517\)                                  & \(0.1458\) & \(0.1501\)                                     &  & \(\mathbf{0.1354}\)                    & \(\mathbf{0.1346}\) & \(\mathbf{0.1487}\)               &  & \(0.4919\)                         & \(0.5172\)          & \(0.4754\) &  & \(0.8997\) & \(1.4341\) & \(2.4387\) &  & \(3.0806\) & \(5.5421\) & \(10.8355\) \\
                \bottomrule
            \end{tabular}
        }
    \end{subtable}\hfill%
    \begin{subtable}[c]{\textwidth}
        \centering
        \caption{LDS and compression wall-time for ResNet9 with CIFAR2 on \Trak{}.}
        \label{subtab:ResNet9_CIFAR2}
        \resizebox{\textwidth}{!}{
            \begin{tabular}{c ccc c ccc c ccc c ccc c ccc c ccc}
                \toprule
                         & \multicolumn{7}{c}{\textbf{Sparsification}} &            & \multicolumn{3}{c}{\textbf{Sparse Projection}} &  & \multicolumn{7}{c}{\textbf{\GraSS{}}} &                     & \multicolumn{3}{c}{\textbf{Baseline}}                                                                                                                                                                                                                                                                                             \\\addlinespace\addlinespace[0.2em]
                         & \multicolumn{3}{c}{\(\RM_{k}\)}             &            & \multicolumn{3}{c}{\(\SM_{k}\)}                &  & \multicolumn{3}{c}{\(\SJLT_{k}\)}     &                     & \multicolumn{3}{c}{\(\SJLT_{k} \circ \RM_{4 k_{\max}}\)} &  & \multicolumn{3}{c}{\(\SJLT_{k} \circ \SM_{4 k_{\max}}\)} &                     & \multicolumn{3}{c}{\(\FJLT_{k}\)}                                                                                                                                                  \\\cline{2-4} \cline{6-8} \cline{10-12} \cline{14-16} \cline{18-20} \cline{22-24}\addlinespace[0.2em]
                \(k\)    & \(2048\)                                    & \(4096\)   & \(8192\)                                       &  & \(2048\)                              & \(4096\)            & \(8192\)                                                 &  & \(2048\)                                                 & \(4096\)            & \(8192\)                          &  & \(2048\)   & \(4096\)   & \(8192\)   &  & \(2048\)   & \(4096\)   & \(8192\)   &  & \(2048\)            & \(4096\)    & \(8192\)            \\
                \midrule
                LDS      & \(0.3690\)                                  & \(0.4116\) & \(0.4236\)                                     &  & \(0.3709\)                            & \(0.4020\)          & \(0.4292\)                                               &  & \(0.4131\)                                               & \(\mathbf{0.4499}\) & \(0.4747\)                        &  & \(0.4123\) & \(0.4357\) & \(0.4545\) &  & \(0.4104\) & \(0.4374\) & \(0.4581\) &  & \(\mathbf{0.4157}\) & \(0.4497\)  & \(\mathbf{0.4753}\) \\
                Time (s) & \(\mathbf{0.1026}\)                         & \(0.1074\) & \(0.1296\)                                     &  & \(0.1032\)                            & \(\mathbf{0.0879}\) & \(\mathbf{0.1134}\)                                      &  & \(12.3590\)                                              & \(12.2393\)         & \(17.4836\)                       &  & \(0.3652\) & \(0.3648\) & \(0.3993\) &  & \(0.3054\) & \(0.2954\) & \(0.2911\) &  & \(31.5491\)         & \(48.1669\) & \(81.9322\)         \\
                \bottomrule
            \end{tabular}
        }
    \end{subtable}\hfill%
    \begin{subtable}[c]{\textwidth}
        \centering
        \caption{LDS and compression wall-time for MusicTransformer with MAESTRO on \Trak{}.}
        \label{subtab:MusicTransformer_MAESTRO}
        \resizebox{\textwidth}{!}{
            \begin{tabular}{c ccc c ccc c ccc c ccc c ccc c ccc}
                \toprule
                         & \multicolumn{7}{c}{\textbf{Sparsification}} &            & \multicolumn{3}{c}{\textbf{Sparse Projection}} &  & \multicolumn{7}{c}{\textbf{\GraSS{}}} &                     & \multicolumn{3}{c}{\textbf{Baseline}}                                                                                                                                                                                                                                                                                          \\\addlinespace\addlinespace[0.2em]
                         & \multicolumn{3}{c}{\(\RM_{k}\)}             &            & \multicolumn{3}{c}{\(\SM_{k}\)}                &  & \multicolumn{3}{c}{\(\SJLT_{k}\)}     &                     & \multicolumn{3}{c}{\(\SJLT_{k} \circ \RM_{4 k_{\max }}\)} &  & \multicolumn{3}{c}{\(\SJLT_{k} \circ \SM_{4 k_{\max }}\)} &             & \multicolumn{3}{c}{\(\FJLT_{k}\)}                                                                                                                                                   \\\cline{2-4} \cline{6-8} \cline{10-12} \cline{14-16} \cline{18-20} \cline{22-24}\addlinespace[0.2em]
                \(k\)    & \(2048\)                                    & \(4096\)   & \(8192\)                                       &  & \(2048\)                              & \(4096\)            & \(8192\)                                                   &  & \(2048\)                                                   & \(4096\)    & \(8192\)                          &  & \(2048\)   & \(4096\)   & \(8192\)   &  & \(2048\)   & \(4096\)   & \(8192\)   &  & \(2048\)     & \(4096\)            & \(8192\)            \\
                \midrule
                LDS      & \(0.2773\)                                  & \(0.2857\) & \(0.3194\)                                     &  & \(0.2662\)                            & \(0.3273\)          & \(0.3733\)                                                 &  & \(\mathbf{0.3062}\)                                        & \(0.3533\)  & \(0.3861\)                        &  & \(0.2826\) & \(0.3378\) & \(0.3755\) &  & \(0.2539\) & \(0.3283\) & \(0.3657\) &  & \(0.2907\)   & \(\mathbf{0.3585}\) & \(\mathbf{0.4011}\) \\
                Time (s) & \(0.5341\)                                  & \(0.5067\) & \(0.5179\)                                     &  & \(\mathbf{0.3800}\)                   & \(\mathbf{0.3971}\) & \(\mathbf{0.4345}\)                                        &  & \(21.6460\)                                                & \(21.1881\) & \(21.3192\)                       &  & \(0.7620\) & \(0.7532\) & \(0.7433\) &  & \(0.7487\) & \(0.7507\) & \(0.7495\) &  & \(100.8136\) & \(156.0613\)        & \(269.9093\)        \\
                \bottomrule
            \end{tabular}
        }
    \end{subtable}\hfill%
    \begin{subtable}[c]{\textwidth}
        \centering
        \caption{LDS and compression wall-time for GPT2-small with WikiText on (block-diagonal FIM) influence function.}
        \label{subtab:GPT2_WikiText}
        \resizebox{\textwidth}{!}{
            \begin{tabular}{c ccc c ccc c ccc c ccc c ccc c ccc}
                \toprule
                         & \multicolumn{7}{c}{\textbf{Sparsification}}                              &                     & \multicolumn{3}{c}{\textbf{Sparse Projection}}                           &  & \multicolumn{7}{c}{\textbf{\FactGraSS{}}}                                  &            & \multicolumn{3}{c}{\textbf{\LoGra{}} (\textbf{Baseline})}                                                                                                                                                                                                                                                                                                                                                                              \\\addlinespace\addlinespace[0.2em]
                         & \multicolumn{3}{c}{\(\RM_{k_l^{\text{in} } \otimes k_l^{\text{out} }}\)} &                     & \multicolumn{3}{c}{\(\SM_{k_l^{\text{in} } \otimes k_l^{\text{out} }}\)} &  & \multicolumn{3}{c}{\(\SJLT_{k_l^{\text{in} } \otimes k_l^{\text{out} }}\)} &            & \multicolumn{3}{c}{\(\SJLT_{k_l} \circ \RM_{2 k_l^{\text{in} } \otimes 2 k_l^{\text{out} }}\)} &  & \multicolumn{3}{c}{\(\SJLT_{k_l} \circ \SM_{2k_l^{\text{in} } \otimes 2k_l^{\text{out} }}\)} &                     & \multicolumn{3}{c}{\(\Gauss_{k_l^{\text{in} } \otimes k_l^{\text{out} } }\)}                                                                                                                                  \\\cline{2-4} \cline{6-8} \cline{10-12} \cline{14-16} \cline{18-20} \cline{22-24}\addlinespace[0.2em]
                \(k_l\)  & \(256\)                                                                  & \(1024\)            & \(4096\)                                                                 &  & \(256\)                                                                    & \(1024\)   & \(4096\)                                                                                       &  & \(256\)                                                                                      & \(1024\)            & \(4096\)                                                                     &  & \(256\)    & \(1024\)   & \(4096\)   &  & \(256\)    & \(1024\)   & \(4096\)   &  & \(256\)     & \(1024\)    & \(4096\)    \\
                \midrule
                LDS      & \(0.1034\)                                                               & \(0.1479\)          & \(\mathbf{0.2391}\)                                                      &  & \(0.0997\)                                                                 & \(0.1617\) & \(0.2267\)                                                                                     &  & \(\mathbf{0.1240}\)                                                                          & \(\mathbf{0.1897}\) & \(0.2389\)                                                                   &  & \(0.1126\) & \(0.1784\) & \(0.2360\) &  & \(0.1102\) & \(0.1860\) & \(0.2380\) &  & \(0.1188\)  & \(0.1818\)  & \(0.2338\)  \\
                Time (s) & \(\mathbf{5.4933}\)                                                      & \(\mathbf{5.3643}\) & \(\mathbf{5.6385}\)                                                      &  & \(5.8603\)                                                                 & \(6.0436\) & \(5.8272\)                                                                                     &  & \(132.5404\)                                                                                 & \(133.4029\)        & \(136.5163\)                                                                 &  & \(6.5790\) & \(7.4161\) & \(6.3075\) &  & \(7.3443\) & \(8.5750\) & \(6.3330\) &  & \(20.4839\) & \(20.9835\) & \(22.2157\) \\
                \bottomrule
            \end{tabular}
        }
    \end{subtable}
\end{table}

\subsection{Scaling up to billion-size model}\label{subsec:scaling-up-to-billion-size-model}
To evaluate the practical utility of \FactGraSS{} in attributing billion-scale models and datasets, we consider Llama-3.1-8B-Instruct~\citep{meta2024llama3} with a random 1B-token subset of the OpenWebText dataset~\citep{Gokaslan2019OpenWeb}, and apply \FactGraSS{} (specifically, \(\SJLT_{k_l} \circ \RM_{2 k_l^{\text{in} } \otimes 2 k_l^{\text{out} }}\)) with layer-wise block-diagonal FIM influence function for linear layers. The experiment is conducted with one \texttt{NVIDIA H200 GPU} with \texttt{96 GB} of memory, and more details can be found in \Cref{adxsubsec:details-of-qualitative-analysis}.

\paragraph{Efficiency.}
We measure the efficiency through the throughputs for \FactGraSS{} and \LoGra{}. \Cref{tab:Llama-8B-throughput} shows the throughput of
\begin{enumerate*}[label=\arabic*.)]
    \item \textbf{compress steps}: compute the projected gradients from inputs and gradients of pre-activation, and the overall
    \item \textbf{cache stage}: compute and save the projected gradients.
\end{enumerate*}

\begin{wraptable}[7]{r}{0.7\textwidth}
    \centering
    \vspace{-1\intextsep}
    \caption{Throughput (tokens per second) for Llama-3.1-8B-Instruct.}
    \label{tab:Llama-8B-throughput}
    \resizebox{0.7\textwidth}{!}{
        \begin{tabular}{lccc c ccc}
            \toprule
                         & \multicolumn{3}{c}{\textbf{Compress}} &                     & \multicolumn{3}{c}{\textbf{Cache}}                                                                   \\
            \cline{2-4} \cline{6-8}\addlinespace[0.2em]
            \(k_l\)      & \(256\)                               & \(1024\)            & \(4096\)                           &  & \(256\)            & \(1024\)           & \(4096\)           \\
            \midrule
            \LoGra{}     & \(27,292\)                            & \(27,255\)          & \(26,863\)                         &  & \(7,307\)          & \(7,478\)          & \(7,367\)          \\
            \FactGraSS{} & \(\mathbf{72,218}\)                   & \(\mathbf{72,684}\) & \(\mathbf{73,811}\)                &  & \(\mathbf{8,584}\) & \(\mathbf{8,594}\) & \(\mathbf{8,681}\) \\
            \bottomrule
        \end{tabular}
    }
\end{wraptable}

We see that \FactGraSS{} significantly improves compute efficiency compared to the previous SOTA, \LoGra{}. In terms of compression steps, we achieve a \(160\%\) faster throughput compared to \LoGra{}, which subsequently improves the overall caching throughput by around \(17\%\). We note that the memory usages are similar in both cases: we set the batch to be \(7\) that maximizes the usage of memory bandwidth for both \LoGra{} and \FactGraSS{}.

\begin{figure}[thpb]
    \centering
    \begin{minipage}{0.49\textwidth}
        \begin{minipage}[c]{0.06\textwidth}
            \includegraphics[height=1em]{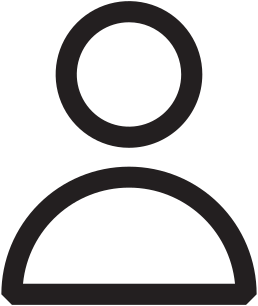}
        \end{minipage}%
        \hfill
        \begin{minipage}[c]{0.93\textwidth}
            {\tiny\fontfamily{lmss}\selectfont
                \begin{tcolorbox}[colback=gray!15,colframe=gray!15,arc=5pt,boxrule=0pt,left=5pt,right=5pt,top=2pt,bottom=2pt]
                    To improve data privacy,
                \end{tcolorbox}
            }
        \end{minipage}

        {\tiny\fontfamily{lmss}\selectfont
        \begin{tcolorbox}[colback=lightgreen,colframe=lightgreen,arc=5pt,boxrule=0pt,left=5pt,right=5pt,top=2pt,bottom=2pt]
            \emph{To improve data privacy,} the European Union has implemented the General Data Protection Regulation (GDPR). \textbf{...}

            \textbf{Data Protection Principles}

            The GDPR sets out six data protection principles\textbf{...}
            \begin{itemize}[leftmargin=*, topsep=0pt, noitemsep]
                \item \textbf{\textcolor{dreamcolor}{Lawfulness}, fairness, and transparency}: Businesses must process \textcolor{dreamcolor}{personal data} in a way that is \textcolor{dreamcolor}{lawful}, fair, and transparent. \textbf{...}
                \item \textbf{Storage limitation}: Businesses must not \textcolor{dreamcolor}{store personal data} for longer than necessary. \textbf{...}
            \end{itemize}

            \textbf{Data Subject Rights}

            The GDPR gives individuals a range of rights when it comes to their \textcolor{dreamcolor}{personal data}. These rights include:
            \begin{itemize}[leftmargin=*, topsep=0pt, noitemsep]
                \item \textbf{Right to access}: Individuals have the \textcolor{dreamcolor}{right to access} their \textcolor{dreamcolor}{personal data} and obtain information about how it is being processed. \textbf{...}
                \item \textbf{Right to erasure}: Individuals have the right to have their \textcolor{dreamcolor}{personal data deleted} if it is no longer necessary for the purposes for which it was collected. \textbf{...} \vspace{-0.8em}\vfill\hfill\includegraphics[height=2em]{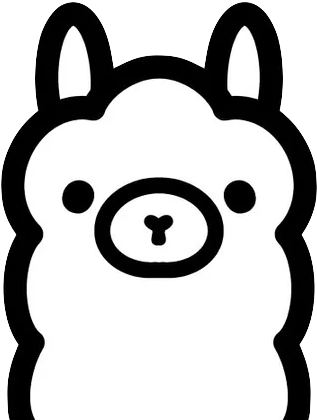}
            \end{itemize}
        \end{tcolorbox}
        }
    \end{minipage}%
    \hfill
    \begin{minipage}{0.49\textwidth}
        {\tiny\fontfamily{lmss}\selectfont
            \begin{tcolorbox}[colback=lightred,colframe=lightred,arc=5pt,boxrule=0pt,left=5pt,right=5pt,top=2pt,bottom=2pt,title={\color{purple}{Influential Data}\hfill\includegraphics[height=1em]{Figures/icon-FactGraSS.png}},fonttitle=\small\bfseries]
                \textbf{...}

                The fact of registration and authorization of users on Sputnik websites via users' account or accounts on social networks indicates acceptance of these rules.

                Users are obliged abide by national and international \textcolor{dreamcolor}{laws}. \textbf{...} The administration has the \textcolor{dreamcolor}{right to delete} comments made in languages other than the language of the majority of the websites \textbf{...}

                \textbf{...}
                \begin{itemize}[leftmargin=*, noitemsep, topsep=0pt]
                    \item \textcolor{dreamcolor}{violates privacy, distributes personal data} of third parties without their consent or \textcolor{dreamcolor}{violates privacy} of correspondence; \textbf{...}
                    \item pursues commercial objectives, contains improper advertising \textcolor{dreamcolor}{unlawful} political advertisement or links to other online resources \textbf{...}
                \end{itemize}
                The administration has the \textcolor{dreamcolor}{right to block a user's access} to the page or \textcolor{dreamcolor}{delete a user's account} without notice if the user is in violation of these rules or if behavior indicating said violation is detected.

                If the moderators deem it possible to \textcolor{dreamcolor}{restore the account/unlock access}, it will be done. In the case of repeated \textcolor{dreamcolor}{violations of the rules} above resulting in a second \textcolor{dreamcolor}{block of a user account, access cannot be restored}. \textbf{...}
            \end{tcolorbox}
        }
    \end{minipage}
    \caption{Qualitative accuracy of data attribution with \FactGraSS{} on Llama-3.1-8B-Instruct.}
    \label{fig:Llama3-8B-quantitative}
\end{figure}

\paragraph{Qualitative Accuracy.}
We next assess the qualitative alignment between the outputs generated by LLMs and the influential data identified by \FactGraSS{} with \(k_l = 4096\). Since naive influence functions often highlight outlier data (e.g., error messages, ASCII codes, or repetitive words) with disproportionately high gradient norms~\citep{choe2024your}, we filter these cases and select the most contextually relevant samples from the top-\(10\) influential data identified by \FactGraSS{}. A representative example is presented in \Cref{fig:Llama3-8B-quantitative}. Given the simple prompt, ``{\fontfamily{lmss}\selectfont To improve data privacy,}’’ \FactGraSS{} identifies a paragraph discussing journalist jailings, including references to privacy policies on various news websites. This content closely aligns with the generated outputs from the model, demonstrating the qualitative accuracy of \FactGraSS{} in capturing relevant data influences.

%% file: 5_conclusion.tex
\section{Conclusion}\label{sec:conclusion}
In this paper, we proposed \GraSS{}, a novel gradient compression algorithm that leverages the inherent sparsity of per-sample gradients to reduce memory and computational overhead significantly. Building on this, we introduced \FactGraSS{}, a variant \GraSS{} that further exploits the gradient structure of linear layers, achieving substantial practical speedups by avoiding ever materializing the full gradient that is both theoretically and practically faster than the previous SOTA baselines.

Our extensive experiments demonstrate that \GraSS{} and \FactGraSS{} consistently outperform existing approaches in both efficiency and scalability, particularly on billion-scale language models.

%% file: acknowledgement.tex
\section*{Acknowledgment}
This work is supported in part by a NAIRR Pilot grant NAIRR240134. WT is partially supported by NSF DMS grant No.\ 2412853, and HZ was partially supported by an NSF IIS grant No.\ 2416897, an NSF CAREER Award No.\ 2442290 and a Google Research Scholar Award. We also thank Xueshen Liu from the University of Michigan for his invaluable discussion in improving the efficiency of our library’s practical implementation. The views and conclusions expressed herein are those of the authors and do not necessarily reflect the official policies or positions of the supporting companies or government agencies.

%% file: 6_appendix.tex
\section{Omitted details from \texorpdfstring{\Cref{sec:preliminary}}{Section 2}}\label{adxsec:preliminary}
\subsection{Related work}\label{adxsubsec:related-work}
This section provides an in-depth discussion of gradient-based data attribution methods and their computational advancements, offering a holistic view of the field. For a more complete overview of other data attribution methods and their applications, we refer interested readers to the recent survey~\citep{deng2025survey}.

\subsubsection{Gradient-based data attribution}
Gradient-based data attribution methods represent a major category for quantifying the impact of individual training samples on a model's behavior~\citep{deng2025survey}. These techniques operate under the principle of using the model's local sensitivity—captured by the \emph{per-sample gradient} with respect to model parameters---to infer how a training example contributes to model predictions or loss values.

The most influential and widely studied approach in this category is the \emph{influence function}~\citep{koh2017understanding}. Rooted in robust statistics, the influence function provides a computationally efficient, first-order approximation of the expensive leave-one-out (LOO) retraining scenario. The core idea is to estimate the change in a target function (such as the loss on a test point) by calculating how the removal of a training sample perturbs the final model parameters. This estimation involves the per-sample gradient and the inverse Hessian matrix. Recent advancements, such as \Trak{}~\citep{park2023trak}, have been developed to improve on this classical framework.

Another family of gradient-based techniques called \emph{training dynamic methods} has become popular due to its ability to adapt to complex training scenarios that might break the influence function assumptions. These include TracIn~\citep{pruthi2020estimating}, SGD-Influence~\citep{hara2019data}, and, more recently, data value embedding~\citep{wang2025capturing}. They analyze how training samples dynamically affect model behavior across intermediate checkpoints of the training trajectory, making them well-suited for non-convex settings where the classical influence function approximations may be less accurate.

All gradient-based methods ultimately rely on similar low-level computations—most notably the inverse Hessian–vector product (iHVP)---which makes the influence function a representative example for discussion.

\subsubsection{Scaling up gradient-based data attribution}
The literature on scaling up gradient-based data attribution methods, or more specifically, iHVP computation, is extensive. In \Cref{sec:preliminary}, we provided a detailed discussion of the \Random{} method, including various random projection techniques. Additionally, in \Cref{subsubssec:FactGraSS:exploiting-layer-wise-gradient-factorization-structure}, we briefly covered block-diagonal, layer-wise independence approximations of the Fisher information matrix (FIM), which together represent the most popular state-of-the-art in efficient approximations of iHVP. Here, we expand on this discussion by providing additional pointers to the broader literature on random projection and sketching for gradient compression, as well as alternative approaches to scaling up iHVP. This context helps clarify the positioning of our proposed methods within the larger landscape of scalable gradient-based data attribution research, offering a more complete understanding of the field.

\paragraph{Sketching and \Random{}.}
Random projection, or sketching, is a well-studied technique for dimensionality reduction, widely explored in both theoretical computer science~\citep{woodruff2014sketching,mahoney2016lecture} and machine learning~\citep{zhang2018billion,liu2021random,chen2019fast,li2023oporp}. In the context of gradient compression, sketching plays a critical role in distributed training, where the overhead of communicating full gradients can be a major bottleneck~\citep{lin2018dgc,aji2017sparse}. However, direct sketching is often avoided in this context, as it can destroy important gradient information for the exact parameter correspondence of the gradient. Instead, techniques like random dropout, which selectively transmit parts of the gradient while preserving critical information, are more common. This is closely related to the Random Mask approach discussed in \Cref{subsec:effective-parameter-sparsity}.

With the rise of gradient-based attribution methods, gradient compression has also become relevant for data attribution~\citep{schioppa2024efficient,lin2024token,choe2024your}. Among these, \citet{schioppa2024efficient} approaches gradient compression from a theoretical sketching perspective, refining traditional methods like the fast Johnson-Lindenstrauss transform (FJLT)~\citep{ailon2009fast,fandina2023fast} to improve computational efficiency on modern machine learning hardware such as TPUs. Notably, only \citet{choe2024your} specifically considers the structural properties of gradients when designing compression methods, potentially offering more accurate reconstructions with reduced communication cost.

\paragraph{Input-Output Independence.}
Two notable extensions of the block-diagonal approximation for empirical Hessians have emerged recently and can be further integrated with \Random{}. The first, Kronecker-Factored Approximate Curvature (K-FAC)~\citep{martens2015kfac}, leverages the Kronecker-factor structure of linear layers (same as \Cref{eq:gradient-factorization}) by assuming independence between the inputs and the pre-activation gradients. This independent factorization significantly reduces the computational burden of Hessian approximations, as inverse FIM-vector product (iFVP) now only requires two smaller inversions of the factorized matrices. Compared to \LoGra{}, where the projected gradients and FIM are materialized at the end with the FIM of size \(k^2\), K-FAC now only materializes two smaller projected inputs and gradients of the pre-activations and their corresponding covariances. The latter two covariances are of size roughly \(\sqrt{k} \times \sqrt{k}\), further reducing the matrix inversion computation.

Building on this, Eigenvalue-corrected K-FAC (EK-FAC)~\citep{grosse2023studying} refines this approach by correcting the eigenvalues of the factorized covariances, improving approximation quality without compromising efficiency. However, we note that while EK-FAC enhances the accuracy of K-FAC, it does not offer further computational speedups.

\paragraph{Direct iHVP.}
An alternative approach to scaling influence functions involves directly estimating the iHVP without explicitly forming or inverting the full Hessian. Unlike the two-stage methods discussed in \Cref{sec:preliminary}, which approximate the full Hessian first and then compute iHVP for each training sample, this direct approach aims to bypass the costly inversion step, providing a more scalable solution for large-scale models.

One such algorithm is LiSSA~\citep{agarwal2017second}, initially developed for stochastic optimization and later adapted for influence function calculations~\citep{koh2017understanding}. It approximates iHVP through iterative stochastic updates that \emph{only involve} Hessian-vector product, which is efficient to compute. While straightforward, this method requires careful tuning of its hyperparameters to balance accuracy and runtime.

More recently, DataInf~\citep{kwon2024datainf} introduced a less conventional approach by reordering the sequence of matrix operations in the iHVP calculation. This method effectively swaps the expectation and inversion steps, allowing per-sample gradient information to approximate the inverse directly. However, this strategy tightly couples the iHVP estimation with the influence calculation, making it challenging to efficiently scale to large datasets, as it requires full computation for each training and test sample pair.

\subsection{A note on Johnson-Lindenstrauss lemma}\label{adsubsec:a-note-on-Johnson-Lindenstrauss-lemma}
While the Johnson-Lindenstrauss lemma~\citep{johnson1984extensions} ensures that the pair-wise distance, and hence the inner products, between gradients are approximately preserved under random i.i.d.\ projection, if \(P\) is not injective when restricted to the range of \(F_{\hat{\theta}}\), i.e., the column rank is not kept, then \(F_{\hat{\theta}}\) applied to vectors in this subspace can generate significant components orthogonal to the subspace. In this case, the projected FIM \(PF_{\hat{\theta}}P^\top\) implicitly neglects these orthogonal components, which leads directly to approximation errors~\citep{schioppa2022scaling}. Several potential strategies to mitigate this issue have been explored, focusing on how to construct the projection matrix \(P\). One popular strategy is to first approximate the top-\(k\) eigenvectors of \(F_{\hat{\theta}} \) using classical algorithms such as PCA~\citep{choe2024your} and Arnoldi iteration~\citep{schioppa2022scaling,arnoldi1951principle}, then used the found top-\(k\) eigenvectors as the rows of \(P\).

\section{Omitted details from \texorpdfstring{\Cref{sec:experiment}}{Section 4}}\label{adxsec:experiment}
In this section, we provide further experimental details that we omitted in \Cref{sec:experiment}.

\subsection{Details of models, datasets, and computing resources}\label{adxsubsec:details-of-models-datasets-and-computing-resources}
We summarize all the models and datasets used in the experiments in \Cref{tab:experimental-detail}.

\begin{table}[thpb]
    \centering
    \caption{Model details used in the experiments.}
    \label{tab:experimental-detail}
    \resizebox{\textwidth}{!}{
        \begin{tabular}{cc | c c c c c}
            \toprule
            \textbf{Models}       & \textbf{Datasets} {\scriptsize(License)} & \textbf{Task}        & \textbf{Parameter Size} & \textbf{Train Samples} & \textbf{Test Samples} & \textbf{Sequential Length} \\
            \midrule
            MLP                   & MNIST {\scriptsize(CC BY-SA 3.0)}        & Image Classification & \(0.11\)M               & \(5,000\)              & \(500\)               & \(1\)                      \\
            ResNet9               & CIFAR2 {\scriptsize(MIT)}                & Image Classification & \(4.83\)M               & \(5,000\)              & \(500\)               & \(1\)                      \\
            Music Transformer     & MAESTRO {\scriptsize(CC BY-NC-SA 4.0)}   & Music Generation     & \(13.3\)M               & \(5,000\)              & \(178\)               & \(1\)                      \\
            GPT2-small            & WikiText {\scriptsize(CC BY-SA 3.0)}     & Text Generation      & \(124\)M                & \(4,656\)              & \(481\)               & \(512\)                    \\
            Llama-3.1-8B-Instruct & OpenWebText {\scriptsize(CC0-1.0)}       & Text Generation      & \(8\)B                  & \(976,562\)            & NA                    & \(1024\)                   \\
            \bottomrule
        \end{tabular}
    }
\end{table}
All the experiments in quantitative analysis are conducted on \texttt{Intel(R) Xeon(R) Gold 6338 CPU @ 2.00GHz} with a single \texttt{Nvidia A40 GPU} with \texttt{48 GB} memory. On the other hand, the qualitative analysis experiment is conducted on the \texttt{VISTA}\footnote{See \url{https://docs.tacc.utexas.edu/hpc/vista/}.} cluster with one Grace Hopper (GH) node, where each GH node has one \texttt{H200 GPU} with \texttt{96 GB} of HBM3 memory and one \texttt{Grace CPU} with \texttt{116 GB} of LPDDR memory.

\subsection{Details of quantitative analysis}\label{adxsubsec:details-of-quantitative-analysis}
\paragraph{Model Training.}
For MLP, ResNet9, and Music Transformer, we utilize pretrained models from the \texttt{dattri} library~\citep[Appendix C]{deng2024dattri}. For GPT2-small, we fine-tune the model on the WikiText dataset using the AdamW optimizer~\citep{loshchilovdecoupled} with a learning rate of \(5 \times 10^{-5}\) and no weight decay, training for \(3\) epochs.

\paragraph{Linear Datamodeling Score (LDS).}
We measure LDS using 50 data subsets, each containing half of the original training set. For each subset, models are trained independently using the hyperparameters described above. For a more comprehensive explanation of the LDS evaluation, we refer readers to \citet{park2023trak}.

\paragraph{Data Attribution.}
For MLP on MNIST, ResNet9 on CIFAR2, and Music Transformer on MAESTRO, we use \Trak{}~\citep{park2023trak} with \(10\), \(10\), and \(5\) independently trained checkpoints, respectively, as the backbone data attribution algorithm to evaluate different gradient compression methods. For GPT2-small fine-tuned on WikiText, we employ a layer-wise block-diagonal FIM approximation for linear layers as the backbone data attribution method.

We remark that one of the important hyperparameters that requires careful attention is the damping term used for the Hessian/FIM inverse. We pick the damping \(\lambda\) for each setting (each model/dataset/compression method combination) via cross-validation grid search for LDS over \(\lambda \in \{10^{-7}, 10^{-6}, 10^{-5}, 10^{-4}, 10^{-3}, 10^{-2}, 10^{-1}, 1, 10, 10^{2}\}\) on \(10\%\) of the test dataset, and evaluate the overall LDS result on the remaining \(90\%\) of the test dataset.

\subsection{Details of qualitative analysis}\label{adxsubsec:details-of-qualitative-analysis}
\paragraph{Model and Dataset.}
For Llama-3.1-8B-Instruct, we directly load the pretrained model without fine-tuning. As for the attribution dataset, while we do not have access to the massive 15T-token pre-training dataset used by Llama-3.1-8B-Instruct, we anticipate that it will contain most of the OpenWebText dataset due to its high quality and popularity.

\subsection{Practical implementation}
Finally, we provide some remarks on the practical implementation of our proposed algorithms.

\subsubsection{SJLT implementation}\label{adxsubsubsec:SJLT-implementation}
A naive implementation of SJLT is straightforward: we first sample random indices corresponding to each input dimension along with their associated signs, and then perform a \texttt{torch.Tensor.index\_add\_()} operation, which carries out the core computation of SJLT (\Cref{fig:sparse-Rademacher-projection}). This operation implicitly uses atomic addition, which can lead to race conditions and slow down computation when the parallelization is not done carefully and the target compression dimension \(k\) is small relative to the input dimension \(p\).

In contrast, our CUDA kernel implementation adopts a key optimization: we parallelize the computation by dividing the input dimension across different threads. This strategy reduces race conditions caused by atomic additions at each step.

\subsubsection{Selective Mask}\label{adxsubsubsec:Selective-Mask}
We discuss several practical considerations and tips for solving \Cref{eq:selective-mask} in the context of Selective Mask.

\paragraph{Ensuring Exact \(k\).}
Since the sparsity of the mask arises from \(\ell_1\) regularization, it is generally not possible to guarantee that the final \(S^{\ast}\) contains exactly \(k\) active indices (i.e., entries greater than \(0.5\)). A simple workaround is to select the top-\(k\) indices based on their sigmoid values---i.e., adaptively setting the activation threshold to ensure exactly \(k\) active indices. However, this method may yield suboptimal masks if the resulting values are far from binary, potentially degrading performance.

To address this issue, we increase the regularization strength and introduce an inverse-temperature term by replacing \(S\) with \(S/T\), where \(T\) decreases as training progresses. As \(T\) approaches zero, \(\sigma(S/T)\) becomes more binary-like, promoting a ``hard'' mask. That said, empirical results show that careful tuning of the regularization parameter \(\lambda\), combined with top-\(k\) selection, can yield performance comparable to the inverse-temperature approach.

\paragraph{Linear Layer.}
For linear layers, we can derive a \emph{factorized} Selective Mask by decomposing the gradients according to \Cref{eq:gradient-factorization}. Specifically, following the notation established in \Cref{subsubssec:FactGraSS:exploiting-layer-wise-gradient-factorization-structure}, for the \(l^{\text{th}}\) linear layer, we can reformulate \Cref{eq:selective-mask} as:
\[
    \begin{split}
        \argmax_{\substack{S_l^{\text{in}}\in \mathbb{R}^{d_l^{\text{in}}}                                                                                \\ S_l^{\text{out}} \in \mathbb{R}^{d_l^{\text{out}}}}} \mathbb{E}_{z_{\text{test}}} &\left[ \operatorname{corr}\left( ( \langle z_{i,l}^{\text{in}} \otimes z_{i,l}^{\text{out}} , z_{\text{test},l}^{\text{in}} \otimes z_{\text{test},l}^{\text{out}} \rangle )_{i=1}^{n}, ( \langle \hat{z}_{i,l}^{\text{in}} \otimes \hat{z}_{i,l}^{\text{out}} , \hat{z}_{\text{test},l}^{\text{in}} \otimes \hat{z}_{\text{test},l}^{\text{out}} \rangle )_{i=1}^{n} \right)\right] \\
         & \qquad\qquad\qquad\qquad\qquad\qquad\qquad\qquad - \lambda (\Vert \sigma(S_l^{\text{in}}) \Vert _1 + \Vert \sigma(S_l^{\text{out}}) \Vert _1),
    \end{split}
\]
where \(z_{i,l}^{\text{in}} \in \mathbb{R}^{d_l^{\text{in}}}\) denotes the input feature of the \(l^{\text{th}}\) linear layer from \(z_i\), \(z_{i,l}^{\text{out}} \in \mathbb{R}^{d_l^{\text{out}}}\) denotes the gradient of its pre-activation of the \(l^{\text{th}}\) linear layer, and \(\hat{z}_{i,l}^{\text{in}} = \sigma(S_l^{\text{in}}) \odot z_{i,l}^{\text{in}}\), \(\hat{z}_{i,l}^{\text{out}} = \sigma(S_l^{\text{out}}) \odot z_{i,l}^{\text{out}}\) are the (soft-)masked variants. Computationally, we leverage the Kronecker product structure to simplify inner product calculations; for example:
\[
    \langle z_{i,l}^{\text{in}} \otimes z_{i,l}^{\text{out}} , z_{\text{test},l}^{\text{in}} \otimes z_{\text{test},l}^{\text{out}} \rangle
    = \langle z_{i,l}^{\text{in}}, z_{\text{test},l}^{\text{in}} \rangle \cdot \langle z_{i,l}^{\text{out}}, z_{\text{test},l}^{\text{out}}\rangle.
\]
As a result, training with the Selective Mask does not require computing full layer-wise gradients, providing similar computational and memory efficiency as in \FactGraSS{}.